\pdfoutput=1

\documentclass[11pt]{article}

\usepackage{acl}

\usepackage{times}
\usepackage{latexsym}
\usepackage{graphicx}

\usepackage[T1]{fontenc}

\usepackage{listings}

\usepackage[utf8]{inputenc}

\usepackage{microtype}

\definecolor{dkgreen}{rgb}{0,0.6,0}
\definecolor{gray}{rgb}{0.5,0.5,0.5}
\definecolor{mauve}{rgb}{0.58,0,0.82}

\title{Solving Linear Algebra by Program Synthesis}

\author{Iddo Drori \\
  MIT \\
  EECS\\
  \texttt{idrori@mit.edu} \\\And
  Nakul Verma \\
  Columbia University\\
  Department of Computer Science\\
  \texttt{verma@cs.columbia.edu} \\}

\begin{document}
\maketitle
\begin{abstract}
We solve MIT's Linear Algebra 18.06 course and Columbia University's Computational Linear Algebra COMS3251 courses with perfect accuracy by interactive program synthesis. This surprisingly strong result is achieved by turning the course questions into programming tasks and then running the programs to produce the correct answers. We use OpenAI Codex with zero-shot learning, without providing any examples in the prompts, to synthesize code from questions. We quantify the difference between the original question text and the transformed question text that yields a correct answer. Since all COMS3251 questions are not available online the model is not overfitting. We go beyond just generating code for questions with numerical answers by interactively generating code that also results visually pleasing plots as output. Finally, we automatically generate new questions given a few sample questions which may be used as new course content. This work is a significant step forward in solving quantitative math problems and opens the door for solving many university level STEM courses by machine.
\end{abstract}

\section{Introduction}
Language models have vastly improved in recent years, with the advent of large-scale Transformer models such as GPT-3  \cite{brown2020language} that perform well on question answering tasks. However, when it comes to answering quantitative problems such as word problems in mathematics or deduction from formal logic, these models show poor performance achieving accuracies close to random baselines \cite{hendrycks2020measuring}, failing on even the most simple questions such as computing the length of a vector.  

Part of the challenge in finding a solution to quantitative problems is to have access to a working tree-like recursive memory. Quantitative problems often require building arithmetic expression trees that help in mathematical deduction. These kinds of trees are also common in program representation and program synthesis. With this insight, we study the efficacy of solving math problems, specifically problems from introductory level undergraduate Linear Algebra courses, by turning each problem into the task of writing a function or program to solve that question. This is done using OpenAI's Codex \cite{chen2021evaluating}, a foundation model trained on both text and code.

We demonstrate the surprisingly simple yet strong result that foundation models for program synthesis such as OpenAI Codex succeed in synthesizing correct code for solving such quantitative math problems. Surprisingly, we find that Codex not only synthesizes correct code for problems that expect numerical answers, but also generates code for questions that ask to plot solutions. We achieve perfect accuracy in solving undergraduate level Linear Algebra course problems, and validate that our results are not merely overfitting the training data by solving a new course which is not available online, and is therefore unseen during Codex training.

\begin{figure*}[ht!]
    \centering
    \includegraphics[width=\textwidth]{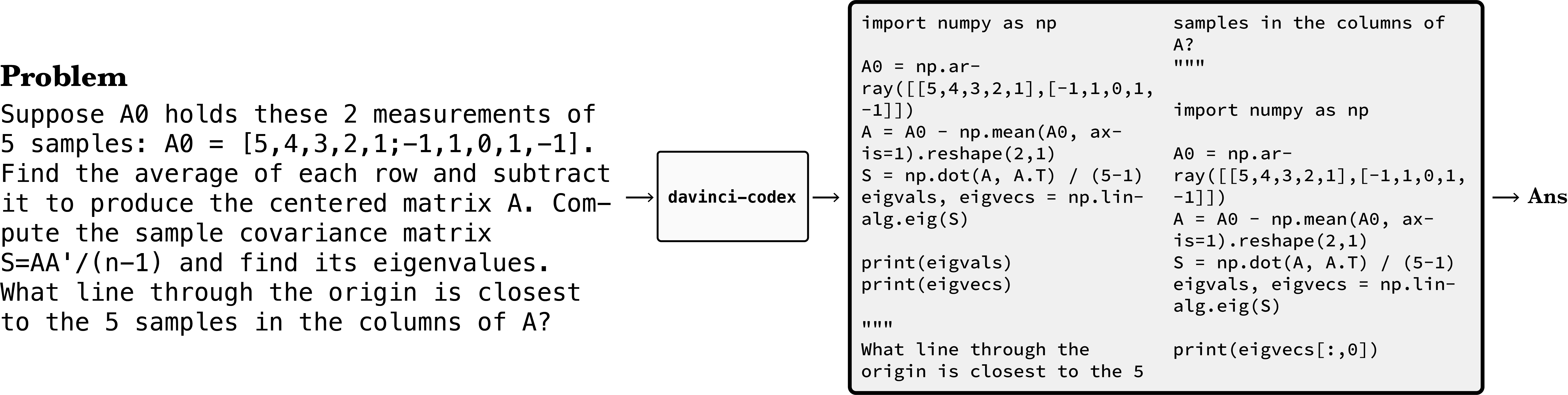}
    \caption{Workflow for solving Linear Algebra questions: (i) Given a question in text, the example shown is Q1 in Ch.\ 7.3 of \citet{strang_lin_alg}, (ii) we run the question through Codex to generate a program, (iii) we execute the program to generate the solution. We transform the question repeating steps (ii) and (iii) until we get it correct.}
    \label{fig:workflow}
\end{figure*}

As an example, consider a moderately involved question from MIT's Linear Algebra course 18.06, Question 1 in Chapter 7.3 of Gilbert Strang's textbook (\citeyear{strang_lin_alg}), as shown in Figure \ref{fig:workflow}. To the best of our knowledge none of the state-of-the-art quantitative reasoning models correctly answers such questions. As shown in Figure \ref{fig:workflow}, given the question as text, we run the question through Codex as is without any modification to generate a program and execute the synthesized program to generate the correct solution. 

\section{Related Work}

There have been several recent works that attempt to improve quantitative reasoning in math problems. MathBERT \cite{peng2021mathbert}, for instance, is a Transformer based pre-trained language model that uses symbol and operator trees as intermediary representations of formulas. 
 
Another line of work has focused on solving math questions from a large database of questions collected from Chinese elementary school math classes. Techniques have included sequence-to-sequence and graph-to-tree Transformers which achieve around $80\%$ on Math23k and MAWPS datasets \cite{koncel2016mawps,li2019modeling,wang2019template,zhang2020graph2tree,li2020graph2tree,wu2020knowledge,qin2020semantically,lan2021mwptoolkit}. Other work \cite{tsai2021sequence} includes knowledge graphs of geometry formulas into sequence-to-tree transformers to improve performance on the geometry section of Math23k dataset, and MWP-BERT, which adds masked fine-tuning to the BERT model using a large corpus of over 100,000 math word problems achieves an impressive $96.2\%$ accuracy on the Math23k  dataset \cite{liang2021mwp} of elementary school math problems.

For solving university level machine learning problems specifically, a recent approach \citep{tran2021solving} uses graph neural networks and Transformers to predict an expression tree from the input question to calculate the answer. This achieves over $95\%$ accuracy on numerical machine learning exercises, which is above human performance; however only works on the specific course it is trained on.

Rather than building a custom-designed solution, our work explores the use of a foundation model such as Codex which is trained on both text and code. Any program may be represented as an abstract syntax tree and many questions may also be represented as expression trees. Bringing the question and answer into a common representation makes it easier to find a correct solution. The advantage of studying this pre-trained model is that it may be applied at scale to many different topics or subjects without additional training. Our work is the first to demonstrate perfect performance of interactively solving linear algebra problems at a university-level difficulty.

\section{Methods}

\begin{figure*}[ht!]
    \centering
    \includegraphics[width=\textwidth]{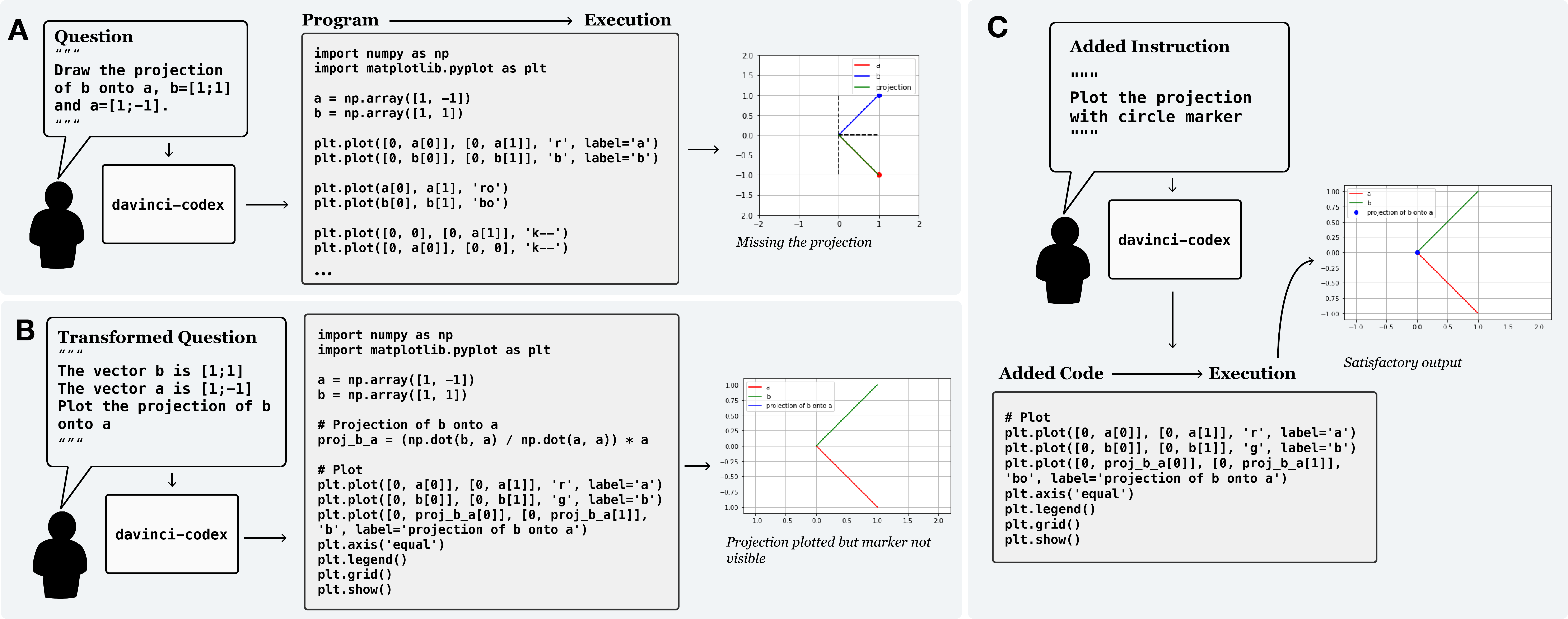}
    \caption{Interactive workflow: (A) We begin with the original question. Codex generates a program which is executed. The result is missing the projection. (B) We transform the question and Codex generates a program again to get the correct answer, though the zero projection vector does not appear on the plot, (C) An additional task to plot the projection vector with a marker so that it is visible results in Codex generating modified code which is executed to yield a correct answer and visually pleasing result.}
    \label{fig:workflow_interactive}
\end{figure*}

Here we describe our dataset, solution generation pipeline, and evaluation methodology. The key components leading to our success are:
\begin{itemize}
    \item Program synthesis: Insight to use a program synthesis to generate a program, that has a built-in tree representation, that produces the solution to the given problem.
    \item Interactive workflow: we interactively work with Codex to produce both the correct result and visually pleasing plots as shown in Figure \ref{fig:workflow_interactive}. We place the question in context by augmenting the question with definitions and information required for the solving the question, rephrase and simplify. See the Appendix for all the original and transformed questions.
\end{itemize}

\subsection{Datasets}
We use (i) problem exercises from Gilbert Strang's \emph{Introduction to Linear Algebra} textbook (\citeyear{strang_lin_alg}), which is used for MIT's Linear Algebra 18.06 course, and (ii) exercises given as homework problems in Columbia's Computational Linear Algebra COMS3251 course, as two challenging real-world university-level datasets. Both courses have multivariable calculus as their prerequisites and are usually taken by second-year EE/CS undergraduate students.
To keep things tractable, we select 3 to 4 random problems from each chapter of the textbook (for MIT 18.06) and from each topic (for COMS3251), resulting in two datasets of 30 questions each for our evaluation. These questions range in difficulty level, and output type (such as a numerical output or drawing a figure with multiple equations). See Figures \ref{fig:workflow}, \ref{fig:workflow_interactive} for examples and the Appendix for a full list of questions.

\subsection{Interactive Workflow}
Our interactive workflow is illustrated in Figure \ref{fig:workflow_interactive}. We begin with the original question from Strang's book (Question 2b, Chapter 4.2), which we feed into Codex that generates a Python program that is then executed. In this example, the result is missing the projection solution. We therefore transform the question to explicitly ask for the projection and have Codex re-generate a program to get the correct answer. The answer also consists of a plot, however the zero projection vector is not visible in the plot since it is a point. We therefore add an additional task which is to plot the projection vector with a marker so that it becomes visible. Codex re-generates the code which is executed to yield both a correct answer and a visually pleasing plot. In all of our experiments we set Codex parameters to be the same fixed default values (using davinci-codex with temperature 0 and response length 200).

\begin{figure}[b!]
    \centering
    \includegraphics[width=0.7\columnwidth]{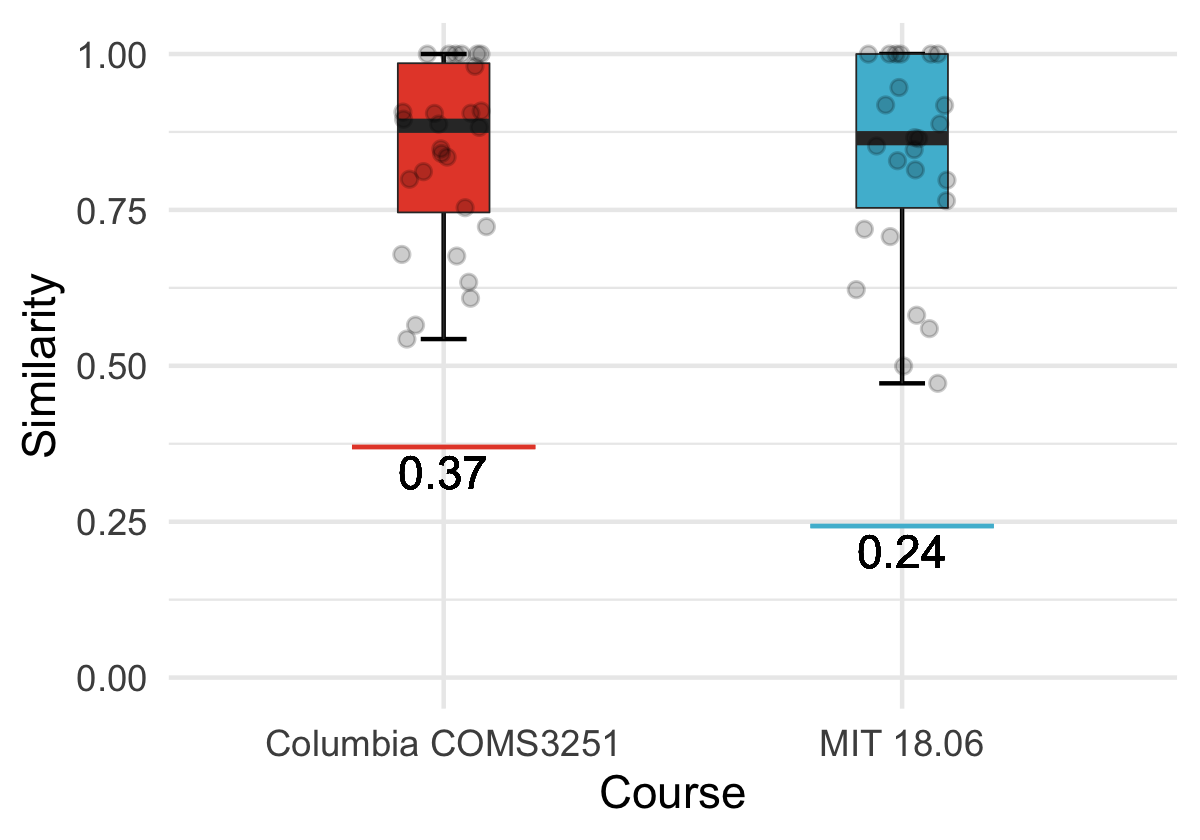}
    \caption{High similarity between the original questions and programming prompts for both COMS3251 and MIT 18.06. Baseline mean pairwise similarity among original questions for both courses shown as a solid horizontal line.}
    \label{fig:course_similarityla}
\end{figure}

\begin{table*}[ht!]
\small
\centering
\begin{tabular}{|l|l|p{6.5cm}|p{6.5cm}|} 
\hline
\textbf{ID} & \textbf{Course} & \textbf{Auto-Generated question} & \textbf{Closest question in the dataset}\\
\hline
1 & MIT 18.06 & Find the eigenvalues and eigenvectors of the matrix $A=[1,1,1;1,2,3;1,3,6]$. & 
Find the eigenvalues of A and B (easy for triangular matrices) and A + B: A = [3,0;1,1], B = [1,1;0,3], A+B = [4,1;1,4].
(Ch 6.1 Q5)
\\
\hline
2 & MIT 18.06 & Find a matrix $A$ such that $A^2$ is not invertible but $A^3$ is invertible. &
Find a matrix that has $A^2$ does not equal 0 but $A^3$ = 0.
(Ch 2.4 Q23b)
\\
\hline
3 & MIT 18.06 & Find a basis for the nullspace of $A = [1,1,1;1,1,1;1,1,1]$. &
Construct a 2 by 2 matrix whose nullspace equals its column space. This is possible. (Ch 3.2 Q20)
\\
\hline
4 & MIT 18.06 & Find a basis for the nullspace of $A$ if the columns of $A$ are unit vectors, all mutually perpendicular. & 
Find A'A if the columns of A are unit vectors, all mutually perpendicular. (Ch 4.1 Q25)
\\ 
\hline
5 & MIT 18.06 & What 2 by 2 matrix $R$ rotates every vector through 90 degrees? & 
What 2 by 2 matrix R rotates every vector through 45 degrees? Example: the vector [1,0] goes to [sqrt(2)/2, sqrt(2)/2].
(Ch 2.1, Q21)
\\
\hline
6 & MIT 18.06 & The complete solution to $Ax = [1;3]$ is $x= [1;0]+c[0;1]$. Find the nullspace of $A$. &
Construct a 2 by 2 matrix whose nullspace equals its column space. This is possible. (Ch 3.2 Q20)
\\
\hline
7 & MIT 18.06 & Find a matrix $A$ that has a negative eigenvalue and is symmetric. & 
Find a symmetric matrix [1,b;b,1] that has a negative eigenvalue. (Ch 6.4, Q9a)
\\
\hline
8 & MIT 18.06 & Find the best plane $C+Dt+Et^2$ to fit $b=[1,2,3,4,5]$ at times $t=0,1,2,3,4$. & 
Find the best line C+Dt to fit b=4,3,-1,0,0 at times t=-2,-1,0,1,2. (Ch 4.3, Q22)
\\
\hline \hline
9 & COMS3251 & Find the eigenvalues of the matrix $A = [1, 2; -2, -3]$. & 
Find the eigenvalues of [-0.2, 0.3; 0.2, -0.3].
\\
\hline
10 & COMS3251 & 
Compute the determinant of the following matrix: $[-1,-2;-2,-4]$ & Compute the determinant of the following matrix:
[3,-4,5;0,-1,-5;5,-4,3] \\
\hline
11 & COMS3251 & 
Find the determinant of the following matrix: $[1,-2,-1;0,2,-3;-4,-5,6]$ & 
Compute the determinant of the following matrix:
[3,-4,5;0,-1,-5;5,-4,3]
\\ 
\hline
12 & COMS3251 & 
Compute an LU decomposition of the matrix $A = [1, 2; -2, -3]$ &
Find an LU decomposition of the following matrix:
[-1,-1,2;2,0,3;-3,2,-1]
\\
\hline
13 & COMS3251 & 
Which of the following matrices is a left inverse to $A=[1,2,-3;-1,-1,0;-2,-3,3]?$
(a) $[-1,0,2;-2,-3,3;-6,-9,9]\;\;\;\;\;\;\;\;\;\;\;\;\;\;\;\;\;\;\;\;\;\;\;\;\;$ 
(b) $[-1,-1,0;0.5,-0.5,0;1/6,-2/6,3/6]\;\;\;\;\;$
(c) $[-1,-2,-3;0.5,-0.5,0;1/6,-4/6,9/6]$ $\;\;\;\;\;\;\;\;\;\;$
(d) None of the above. & 
Compute the inverse of the following matrix:
[-1,-2;-2,0]
\\
\hline
14 & COMS3251 & 
Find a combination of the vectors $[1; 2; 3]$, $[4; 5; 6]$, and $[7; 8; 9]$ that gives the vector $[12; 23]$ & 
Find a combination of the vectors [1; 2; 3], [4; 5; 6], and [7; 8; 9] that give the zero vector.
\\
\hline
15 & COMS3251 & 
What is the dimension of the subspace spanned by the following vectors?
$[1,2,3]$,$[0,1,0]$,$[-1/2,-1/3,1]$ &
What is the dimension of the subspace spanned by the following vectors?
[2,-1/2],[1,1],[4,4]
\\
\hline
16 & COMS3251 & 
Find the projection matrix onto the column space of $A = [1, 2, 3; 4, 5, 6]$ & 
Find the projection matrix onto the column space of A [3, 6, 6; 4, 8, 8].
\\
\hline
\end{tabular}
\caption{New questions generated from MIT Linear Algebra 18.06 questions and Computational Linear Algebra (COMS3251) questions, and the closest question among the existing questions.}
\label{tab:new18.06newCOMS3251}
\end{table*}

\section{Results}
\subsection{Performance Evaluation}
Our dataset includes 30 questions from MIT's 18.06 and 30 questions from Columbia University's COMS3251 and gets perfect accuracy on these courses (see Appendix for detailed input and the solution output for each question in the datasets). In contrast, GPT3 yields 0\% accuracy. We would like to quantify the extent of human effort required for achieving these perfect results. We therefore measure the similarity between the original question text and the final programming prompt that results in a correct answer. As shown in Figure \ref{fig:course_similarityla} we observe highly similar texts. Specifically 90\% median similarity for Columbia University's COMS3251 and 80\% median similarity for MIT's 18.06, computed using the cosine similarity between their language embeddings. This demonstrates that only minor changes are required for turning a question into a program task that results in a correct answer. As a baseline we also include the similarity among the different  original questions in each course, to verify the validity of our metric.

\subsection{Generating New Questions}
We are able to generate new questions with ease. We prompt Codex by a set of $n$ numbered questions on different topics, and synthesize question number $n+1$. Table \ref{tab:new18.06newCOMS3251} shows eight new generated questions for each course.

\subsection{Limitations}
We currently do not handle input drawings or any visual elements as input. Extending our approach to handle such inputs by using a multi-modal text and vision Transformer would help solve many diverse types of mathematical problems. While our methodology works well for numerical outputs and figures, our pipeline doesn't yet handle solutions that require multi-line derivations or proofs. We currently often modify the original question manually to form a question for which Codex returns a program which solves the question correctly making our method interactive. We plan on training a Transformer, such as T5 \cite{T5}, for paraphrazing and performing this step automatically.

\section{Conclusion}
Our work is the first to solve linear algebra problems at a university-level difficulty. Our results open the door for solving other STEM courses, which has the potential to disrupt higher education by: (i) automatically learning all university level STEM courses, (ii) automatically grading course, and (iii) rapidly generating new course content.

\newpage
\clearpage

\bibliography{bibliography}

\begin{thebibliography}{17}
\expandafter\ifx\csname natexlab\endcsname\relax\def\natexlab#1{#1}\fi

\bibitem[{Brown et~al.(2020)Brown, Mann, Ryder, Subbiah, Kaplan, Dhariwal,
  Neelakantan, Shyam, Sastry, Askell et~al.}]{brown2020language}
Tom~B Brown, Benjamin Mann, Nick Ryder, Melanie Subbiah, Jared Kaplan, Prafulla
  Dhariwal, Arvind Neelakantan, Pranav Shyam, Girish Sastry, Amanda Askell,
  et~al. 2020.
\newblock Language models are few-shot learners.
\newblock \emph{arXiv preprint arXiv:2005.14165}.

\bibitem[{Chen et~al.(2021)}]{chen2021evaluating}
Mark Chen et~al. 2021.
\newblock \href {http://arxiv.org/abs/2107.03374} {Evaluating large language
  models trained on code}.

\bibitem[{Hendrycks et~al.(2020)Hendrycks, Burns, Basart, Zou, Mazeika, Song,
  and Steinhardt}]{hendrycks2020measuring}
Dan Hendrycks, Collin Burns, Steven Basart, Andy Zou, Mantas Mazeika, Dawn
  Song, and Jacob Steinhardt. 2020.
\newblock Measuring massive multitask language understanding.
\newblock \emph{arXiv preprint arXiv:2009.03300}.

\bibitem[{Koncel-Kedziorski et~al.(2016)Koncel-Kedziorski, Roy, Amini, Kushman,
  and Hajishirzi}]{koncel2016mawps}
Rik Koncel-Kedziorski, Subhro Roy, Aida Amini, Nate Kushman, and Hannaneh
  Hajishirzi. 2016.
\newblock {MAWPS}: {A} math word problem repository.
\newblock In \emph{Conference of the North American Chapter of the Association
  for Computational Linguistics: Human Language Technologies}, pages
  1152--1157.

\bibitem[{Lan et~al.(2021)Lan, Wang, Zhang, Lan, Dai, Wang, Zhang, and
  Lim}]{lan2021mwptoolkit}
Yihuai Lan, Lei Wang, Qiyuan Zhang, Yunshi Lan, Bing~Tian Dai, Yan Wang,
  Dongxiang Zhang, and Ee-Peng Lim. 2021.
\newblock Mwptoolkit: An open-source framework for deep learning-based math
  word problem solvers.
\newblock \emph{arXiv preprint arXiv:2109.00799}.

\bibitem[{Li et~al.(2019)Li, Wang, Zhang, Wang, Dai, and
  Zhang}]{li2019modeling}
Jierui Li, Lei Wang, Jipeng Zhang, Yan Wang, Bing~Tian Dai, and Dongxiang
  Zhang. 2019.
\newblock Modeling intra-relation in math word problems with different
  functional multi-head attentions.
\newblock In \emph{Proceedings of Annual Meeting of the Association for
  Computational Linguistics}, pages 6162--6167.

\bibitem[{Li et~al.(2020)Li, Wu, Feng, Xu, Xu, and Zhong}]{li2020graph2tree}
Shucheng Li, Lingfei Wu, Shiwei Feng, Fangli Xu, Fengyuan Xu, and Sheng Zhong.
  2020.
\newblock Graph-to-tree neural networks for learning structured input-output
  translation with applications to semantic parsing and math word problem.
\newblock In \emph{Conference on Empirical Methods in Natural Language
  Processing}, pages 2841--2852.

\bibitem[{Liang et~al.(2021)Liang, Zhang, Shao, and Zhang}]{liang2021mwp}
Zhenwen Liang, Jipeng Zhang, Jie Shao, and Xiangliang Zhang. 2021.
\newblock Mwp-bert: A strong baseline for math word problems.
\newblock \emph{arXiv preprint arXiv:2107.13435}.

\bibitem[{Peng et~al.(2021)Peng, Yuan, Gao, and Tang}]{peng2021mathbert}
Shuai Peng, Ke~Yuan, Liangcai Gao, and Zhi Tang. 2021.
\newblock Mathbert: A pre-trained model for mathematical formula understanding.
\newblock \emph{arXiv preprint arXiv:2105.00377}.

\bibitem[{Qin et~al.(2020)Qin, Lin, Liang, Zhang, and
  Lin}]{qin2020semantically}
Jinghui Qin, Lihui Lin, Xiaodan Liang, Rumin Zhang, and Liang Lin. 2020.
\newblock Semantically-aligned universal tree-structured solver for math word
  problems.
\newblock \emph{Conference on Empirical Methods in Natural Language
  Processing}.

\bibitem[{Raffel et~al.(2020)Raffel, Shazeer, Roberts, Lee, Narang, Matena,
  Zhou, Li, and Liu}]{T5}
Colin Raffel, Noam Shazeer, Adam Roberts, Katherine Lee, Sharan Narang, Michael
  Matena, Yanqi Zhou, Wei Li, and Peter~J. Liu. 2020.
\newblock \href {http://arxiv.org/abs/1910.10683} {Exploring the limits of
  transfer learning with a unified text-to-text transformer}.

\bibitem[{Strang(2016)}]{strang_lin_alg}
G.~Strang. 2016.
\newblock \href {https://books.google.com/books?id=efbxjwEACAAJ}
  {\emph{Introduction to Linear Algebra}}.
\newblock Wellesley-Cambridge Press.

\bibitem[{Tran et~al.(2021)Tran, Krishna, Pakuwal, Kafle, Singh, Lynch, and
  Drori}]{tran2021solving}
Sunny Tran, Pranav Krishna, Ishan Pakuwal, Prabhakar Kafle, Nikhil Singh,
  Jayson Lynch, and Iddo Drori. 2021.
\newblock Solving machine learning problems.
\newblock \emph{Asian Conference on Machine Learning}.

\bibitem[{Tsai et~al.(2021)Tsai, Liang, Wang, and Su}]{tsai2021sequence}
Shih-hung Tsai, Chao-Chun Liang, Hsin-Min Wang, and Keh-Yih Su. 2021.
\newblock Sequence to general tree: Knowledge-guided geometry word problem
  solving.
\newblock \emph{arXiv preprint arXiv:2106.00990}.

\bibitem[{Wang et~al.(2019)Wang, Zhang, Zhang, Xu, Gao, Dai, and
  Shen}]{wang2019template}
Lei Wang, Dongxiang Zhang, Jipeng Zhang, Xing Xu, Lianli Gao, Bing~Tian Dai,
  and Heng~Tao Shen. 2019.
\newblock Template-based math word problem solvers with recursive neural
  networks.
\newblock In \emph{Proceedings of the AAAI Conference on Artificial
  Intelligence}, volume~33, pages 7144--7151.

\bibitem[{Wu et~al.(2020)Wu, Zhang, Fu, and Huang}]{wu2020knowledge}
Qinzhuo Wu, Qi~Zhang, Jinlan Fu, and Xuan-Jing Huang. 2020.
\newblock A knowledge-aware sequence-to-tree network for math word problem
  solving.
\newblock In \emph{Conference on Empirical Methods in Natural Language
  Processing}, pages 7137--7146.

\bibitem[{Zhang et~al.(2020)Zhang, Wang, Lee, Bin, Wang, Shao, and
  Lim}]{zhang2020graph2tree}
Jipeng Zhang, Lei Wang, Roy Ka-Wei Lee, Yi~Bin, Yan Wang, Jie Shao, and Ee-Peng
  Lim. 2020.
\newblock Graph-to-tree learning for solving math word problems.
\newblock In \emph{Proceedings of the Annual Meeting of the Association for
  Computational Linguistics}, pages 3928--3937.

\end{thebibliography}
\bibliographystyle{acl_natbib}

\onecolumn
\appendix
\section{Computational Linear Algebra: Columbia University COMS3251}
\label{sec:appendixCOMS3251}

\begin{table}[h]
\small
\centering
\begin{tabular}{|p{3.5cm}|p{11.5cm}|}
\hline
Topic & Matrix Algebra\\
\hline
Original question & Compute the following expression: $([-1,0,2; 0,1,4]'*[-2;1]).([3,1;0,0;-2,-1]*([5,-3]+[-4,2])')$, where ' means transpose, * means matrix product, . means inner product.\\
\hline
Codex input & Compute the following expression: $(\textit{transpose}([-1,0,2; 0,1,4])*[-2;1]).([3,1;0,0;-2,-1]*\textit{transpose}([5,-3]+[-4,2]))$\\
\hline
Codex output & \begin{lstlisting}
import numpy as np

a = np.array([[-1,0,2],[0,1,4]])
b = np.array([[-2],[1]])
c = np.array([[3,1],[0,0],[-2,-1]])
d = np.array([[5],[-3]])
e = np.array([[-4],[2]])

print(np.dot(np.transpose(a),b))
print(np.dot(c,(d+e)))
\end{lstlisting}\\
\hline
Solution & $4$\\
\hline
\end{tabular}
\caption*{COMS3251, Question 1: Original question, Codex input and output, and solution.}
\end{table}

\begin{table}[h]
\small
\centering
\begin{tabular}{|p{3.5cm}|p{11.5cm}|}
\hline
Topic & Vectors, Lengths and Dot products \\
\hline
Original question & 
Compute the squared $L_{2}$ norm of the vector $[1;-4;2;8;-1]$.\\
\hline
Codex input & Original question\\
\hline
Codex output & \begin{lstlisting}
import numpy as np

def compute_squared_L2_norm(vector):
    return np.sum(np.square(vector))

vector = np.array([1, -4, 2, 8, -1])
print(compute_squared_L2_norm(vector))
\end{lstlisting}\\
\hline
Solution & $86$\\
\hline
\end{tabular}
\caption*{COMS3251, Question 2: Original question, Codex input and output, and solution.}
\end{table}

\begin{table}[h]
\small
\centering
\begin{tabular}{|p{3.5cm}|p{11.5cm}|}
\hline
Topic & Solving Linear System of Equations\\
\hline
Original question & Find a solution to the following system of four equations:
     $4x - 2y  +  8z   +   w  =  3,
    -8x + 10y +          3w  = -2,
     3x - 1y  +  10z  +  5w  = -1,
     2x + 2y  +  9z   -  2w  =  8$.\\
\hline
Codex input & Write a program to find a solution to the system of four equations:
     $4x - 2y  +  8z   +   w  =  3,
    -8x + 10y +          3w  + 2 = 0,
     3x - 1y  +  10z  +  5w  + 1 = 0,
     2x + 2y  +  9z   -  2w  =  8$\\
\hline
Codex output & \begin{lstlisting}
from sympy import *

x, y, z, w = symbols('x y z w')

eq1 = 4*x - 2*y + 8*z + w - 3
eq2 = -8*x + 10*y + 3*w + 2
eq3 = 3*x - 1*y + 10*z + 5*w + 1
eq4 = 2*x + 2*y + 9*z - 2*w - 8

solve([eq1, eq2, eq3, eq4], [x, y, z, w])
\end{lstlisting}\\
\hline
Solution & $x=11/4, y=97/44, z=-4/11, w=-15/22$\\
\hline
\end{tabular}
\caption*{COMS3251, Question 3: Original question, Codex input and output, and solution.}
\end{table}

\begin{table}[h]
\small
\centering
\begin{tabular}{|p{3.5cm}|p{11.5cm}|}
\hline
Topic & Solving Linear System of Equations\\
\hline
Original question & 
A mining company has two mines. One day's operation at mine A produces ore that contains 10 metric tons of copper and 120 kilograms of silver, while one day's operation at mine B produces ore containing 6 metric tons of copper and 140 kilograms of silver. Let a=[10;120] and b=[6;140]. Then a and b represent the daily output of mines A and B, respectively. Suppose that the mining company operates mine A for x days and mine B for y days.
Calculate how many number of days each mine should operate in order to produce 72 tons of copper and 1340 kilograms of silver.
\\
\hline
Codex input & 
Let a=[10;120] and b=[6;140].
Write a program to compute positive x and y such that x*a + y*b = [72;1340]
\\
\hline
Codex output & \begin{lstlisting}
a = np.array([10, 120])
b = np.array([6, 140])
c = np.array([72, 1340])

# Solve for x and y in the equation x*a + y*b = c using
# the numpy linear algebra function linalg.solve(a, b)
x, y = np.linalg.solve(np.vstack((a, b)).T, c)
print(""x: "", x)
print(""y: "", y)
\end{lstlisting}\\
\hline
Solution & x=3, y=7\\
\hline
\end{tabular}
\caption*{COMS3251, Question 4: Original question, Codex input and output, and solution.}
\end{table}

\begin{table}[h]
\small
\centering
\begin{tabular}{|p{3.5cm}|p{11.5cm}|}
\hline
Topic & Vectors, Lengths and Dot products\\
\hline
Original question & 
At noon, the minute and the hour hands of an analog clock coincide. What is the angle (in degrees) subtended by the minute and hour hands of an analog clock at 1:15?
\\
\hline
Codex input & 
At noon, the minute and the hour hands of an analog clock coincide.
Write a program to compute the angle (in degrees) subtended by the hands of an analog clock at 1:15
\\
\hline
Codex output & \begin{lstlisting}
def clock_angle(hour, minute):
    hour_angle = (hour * 30) + (minute * 0.5)
    minute_angle = (minute * 6)
    angle = abs(hour_angle - minute_angle)
    return angle

print(clock_angle(1, 15))
\end{lstlisting}\\
\hline
Solution & 
52.5
\\
\hline
\end{tabular}
\caption*{COMS3251, Question 5: Original question, Codex input and output, and solution.}
\end{table}

\begin{table}[h]
\small
\centering
\begin{tabular}{|p{3.5cm}|p{11.5cm}|}
\hline
Topic & Matrix Algebra\\
\hline
Original question & 
For what matrix R, the following matrix equation is satisfied?
R*[1,0,0,0,0;1,1,0,0,0;1,2,1,0,0;1,3,3,1,0;1,4,6,4,1]=[1,0,0,0,0;0,1,0,0,0;
0,1,1,0,0;0,1,2,1,0;0,1,3,3,1]
\\
\hline
Codex input & 
A = np.matrix([[1,0,0,0,0],[1,1,0,0,0],[1,2,1,0,0],[1,3,3,1,0],[1,4,6,4,1]])
B = np.matrix([[1,0,0,0,0],[0,1,0,0,0],[0,1,1,0,0],[0,1,2,1,0],[0,1,3,3,1]])
""""""
Solve X*A = B for X
""""""
\\
\hline
Codex output & \begin{lstlisting}
# Solve X*A = B for X
X = np.linalg.solve(A,B)
print(X)
\end{lstlisting}\\
\hline
Solution & 
[1,0,0,0,0;-1,1,0,0,0;0,-1,1,0,0;0,0,-1,1,0;0,0,0,-1,1]
\\
\hline
\end{tabular}
\caption*{COMS3251, Question 6: Original question, Codex input and output, and solution.}
\end{table}

\begin{table}[h]
\small
\centering
\begin{tabular}{|p{3.5cm}|p{11.5cm}|}
\hline
Topic & Rank, Basis and Dimension\\
\hline
Original question & 
Given a d-dimensional non-zero vector v, compute the rank of the matrix v*v'
\\
\hline
Codex input & 
Given a d-dimensional non-zero vector v, write a program to compute the rank of the matrix v*transpose(v)
\\
\hline
Codex output & \begin{lstlisting}
import numpy as np

def rank(v):
    return np.linalg.matrix_rank(np.dot(v, v.T))
\end{lstlisting}\\
\hline
Solution & 1\\
\hline
\end{tabular}
\caption*{COMS3251, Question 7: Original question, Codex input and output, and solution.}
\end{table}

\begin{table}[h]
\small
\centering
\begin{tabular}{|p{3.5cm}|p{11.5cm}|}
\hline
Topic & Four Fundamental Subspaces of a Matrix\\
\hline
Original question & 
Compute the dimension of the null space of the following matrix. [1,2,0,-1;-2,-3,4,5;2,4,0,-2]
\\
\hline
Codex input & 
Write a program that finds the finds the dimension the null space of the matrix [1,2,0,-1;-2,-3,4,5;2,4,0,-2]. Use scipy.linalg.null\_space. Get the shape[1] of the null space.
\\
\hline
Codex output & \begin{lstlisting}
import numpy as np
from scipy import linalg

A = np.array([[1,2,0,-1],[-2,-3,4,5],[2,4,0,-2]])

print(linalg.null_space(A))

print(linalg.null_space(A).shape[1])
\end{lstlisting}\\
\hline
Solution & 
2\\
\hline
\end{tabular}
\caption*{COMS3251, Question 8: Original question, Codex input and output, and solution.}
\end{table}

\begin{table}[h]
\small
\centering
\begin{tabular}{|p{3.5cm}|p{11.5cm}|}
\hline
Topic & Four Fundamental Subspaces of a Matrix\\
\hline
Original question & 
Compute the dimension of the left null space of the following matrix. [1,2;2,4;3,6;4,8]
\\
\hline
Codex input & 
Compute the dimension of the left null space of the matrix [1,2;2,4;3,6;4,8]
\\
\hline
Codex output & \begin{lstlisting}
import numpy as np
A = np.array([[1,2],[2,4],[3,6],[4,8]])
print(np.linalg.matrix_rank(A))
\end{lstlisting}\\
\hline
Solution & 
3
\\
\hline
\end{tabular}
\caption*{COMS3251, Question 9: Original question, Codex input and output, and solution.}
\end{table}

\begin{table}[h]
\small
\centering
\begin{tabular}{|p{3.5cm}|p{11.5cm}|}
\hline
Topic & Orthogonality and Projections\\
\hline
Original question &
Compute the orthogonal projection of the vector [-6;4] onto the line spanned by the vector [3;2]
\\
\hline
Codex input & 
Compute the orthogonal projection of the vector [-6;4] onto the line spanned by the vector [3;2]
\\
\hline
Codex output & \begin{lstlisting}
import numpy as np

# Define the two vectors
v1 = np.array([3, 2])
v2 = np.array([-6, 4])

# Compute the orthogonal projection of v2 onto v1 and print it out
proj = (np.dot(v1, v2) / np.dot(v1, v1)) * v1
print(proj)
\end{lstlisting}\\
\hline
Solution & 
[-30/13;-20/13]
\\
\hline
\end{tabular}
\caption*{COMS3251, Question 10: Original question, Codex input and output, and solution.}
\end{table}

\begin{table}[h]
\small
\centering
\begin{tabular}{|p{3.5cm}|p{11.5cm}|}
\hline
Topic & Least Squares Approximation\\
\hline
Original question & 
Find the least squares solution to the following matrix equation.
[0,1;1,1;2,1]*x=[6;0;0]
\\
\hline
Codex input & 
Write a program that finds the least squares solution to the matrix equation [0,1;1,1;2,1]*x = [6;0;0]
\\
\hline
Codex output & \begin{lstlisting}
import numpy as np
A = np.array([[0,1],[1,1],[2,1]])
b = np.array([6,0,0])
x = np.linalg.solve(A,b)
print(x)
\end{lstlisting}\\
\hline
Solution & 
[-3;5]
\\
\hline
\end{tabular}
\caption*{COMS3251, Question 11: Original question, Codex input and output, and solution.}
\end{table}

\begin{table}[h]
\small
\centering
\begin{tabular}{|p{3.5cm}|p{11.5cm}|}
\hline
Topic & Eigenvectors and Eigenvalues\\
\hline
Original question & 
Compute the eigenvalue associated with with the eigenvector [-684/721;228/721] of the matrix [-6,3;4,5]
\\
\hline
Codex input & 
Write a program that finds the eigenvalue associated with with the eigenvector [-684/721;228/721] of the matrix [-6,3;4,5]
\\
\hline
Codex output & \begin{lstlisting}
import numpy as np
A = np.array([[-6,3],[4,5]])
eigenvalues, eigenvectors = np.linalg.eig(A)
print(eigenvalues)
print(eigenvectors)
print(eigenvalues[0]*eigenvectors[:,0])
print(eigenvalues[1]*eigenvectors[:,1])
\end{lstlisting}\\
\hline
Solution &
-7
\\
\hline
\end{tabular}
\caption*{COMS3251, Question 12: Original question, Codex input and output, and solution.}
\end{table}

\begin{table}[h]
\small
\centering
\begin{tabular}{|p{3.5cm}|p{11.5cm}|}
\hline
Topic & Matrix Algebra\\
\hline
Original question & 
Compute the inverse of the following matrix:
[-1,-2;-2,0]
\\
\hline
Codex input & 
Compute the inverse of the following matrix:
[-1,-2;-2,0]
\\
\hline
Codex output & \begin{lstlisting}
import numpy as np
A = np.array([[-1,-2],[-2,0]])
print(np.linalg.inv(A))
\end{lstlisting}\\
\hline
Solution & 
[0,-1/2;-1/2,1/4]
\\
\hline
\end{tabular}
\caption*{COMS3251, Question 13: Original question, Codex input and output, and solution.}
\end{table}

\begin{table}[h]
\small
\centering
\begin{tabular}{|p{3.5cm}|p{11.5cm}|}
\hline
Topic & Determinants\\
\hline
Original question & 
Compute the determinant of the following matrix:
[3,-4,5;0,-1,-5;5,-4,3]
\\
\hline
Codex input & 
Compute the determinant of the following matrix:
[3,-4,5;0,-1,-5;5,-4,3]
\\
\hline
Codex output & \begin{lstlisting}
import numpy as np
matrix = np.array([[3,-4,5],[0,-1,-5],[5,-4,3]])
print(np.linalg.det(matrix))
\end{lstlisting}\\
\hline
Solution & 
56
\\
\hline
\end{tabular}
\caption*{COMS3251, Question 14: Original question, Codex input and output, and solution.}
\end{table}

\begin{table}[h]
\small
\centering
\begin{tabular}{|p{3.5cm}|p{11.5cm}|}
\hline
Topic & Special Matrices\\
\hline
Original question & 
Find the real numbers a, b and c so that the following matrix is symmetric.
[0,a+b,c+2;a,2,c;4,a+b,4]
\\
\hline
Codex input & 
Use sympy to find a, b, c so that [0,a+b,c+2;a,2,c;4,a+b,4] = transpose([0,a+b,c+2;a,2,c;4,a+b,4])
\\
\hline
Codex output & \begin{lstlisting}
import sympy

a, b, c = sympy.symbols('a b c')
A = sympy.Matrix([[0, a + b, c + 2], [a, 2, c], [4, a + b, 4]])
B = A.transpose()
print(sympy.solve([A[i] - B[i] for i in range(3)], (a, b, c)))
\end{lstlisting}\\
\hline
Solution & 
a=2,b=0,c=2
\\
\hline
\end{tabular}
\caption*{COMS3251, Question 15: Original question, Codex input and output, and solution.}
\end{table}

\begin{table}[h]
\small
\centering
\begin{tabular}{|p{3.5cm}|p{11.5cm}|}
\hline
Topic & Matrix Decomposition\\
\hline
Original question & 
Find an LU decomposition of the following matrix:
[-1,-1,2;2,0,3;-3,2,-1]
\\
\hline
Codex input & 
Find an LU decomposition of the following matrix:
[-1,-1,2;2,0,3;-3,2,-1]
\\
\hline
Codex output & \begin{lstlisting}
import numpy as np
A = np.array([[-1,-1,2],[2,0,3],[-3,2,-1]])
L = np.eye(3)
U = A
for i in range(3):
    for j in range(i+1,3):
        L[j,i] = U[j,i]/U[i,i]
        U[j,:] = U[j,:] - L[j,i]*U[i,:]
print(L)
print(U)
\end{lstlisting}\\
\hline
Solution &
[1,0,0;-2,1,0;3,-5/2,1]*[-1,-1,2;0,-2,7;0,0,21/2]
\\
\hline
\end{tabular}
\caption*{COMS3251, Question 16: Original question, Codex input and output, and solution.}
\end{table}

\begin{table}[h]
\small
\centering
\begin{tabular}{|p{3.5cm}|p{11.5cm}|}
\hline
Topic & Matrix Decomposition\\
\hline
Original question & 
Find an QR decomposition of the following matrix:
[1,0,2;0,2,0;0,-1,1]
\\
\hline
Codex input &
Find an QR decomposition of the following matrix:
[1,0,2;0,2,0;0,-1,1]
\\
\hline
Codex output & \begin{lstlisting}
import numpy as np
A = np.array([[1,0,2],[0,2,0],[0,-1,1]])
Q,R = np.linalg.qr(A)
print(Q)
print(R)
\end{lstlisting}\\
\hline
Solution & 
[1,0,0;0,2/sqrt(5),1/sqrt(5);0,-1/sqrt(5),2/sqrt(5)]*[1,0,2;0,sqrt(5),-1/sqrt(5);0,0,2/sqrt(5)]
\\
\hline
\end{tabular}
\caption*{COMS3251, Question 17: Original question, Codex input and output, and solution.}
\end{table}

\begin{table}[h]
\small
\centering
\begin{tabular}{|p{3.5cm}|p{11.5cm}|}
\hline
Topic & Diagonalization and Eigenvectors/Eigenvalues\\
\hline
Original question & 
Diagonalize the following matrix:
[2,-3,0;0,-1,0;1,3,1]
\\
\hline
Codex input & 
Write a program that diagonalizes the matrix [2,-3,0;0,-1,0;1,3,1]
\\
\hline
Codex output & \begin{lstlisting}
import numpy as np
A = np.array([[2,-3,0],[0,-1,0],[1,3,1]])
print(A)
eig_val, eig_vec = np.linalg.eig(A)
print(eig_val)
print(eig_vec)
D = np.diag(eig_val)
print(D)
V = eig_vec
print(V)
V_inv = np.linalg.inv(V)
print(V_inv)
print(np.dot(V,np.dot(D,V_inv)))
\end{lstlisting}\\
\hline
Solution & 
[-1,0,1;-1,0,0;2,1,1]*[-1,0,0;0,1,0;0,0,2]*[0,-1,0;-1,3,1;1,-1,0]
\\
\hline
\end{tabular}
\caption*{COMS3251, Question 18: Original question, Codex input and output, and solution.}
\end{table}

\begin{table}[h]
\small
\centering
\begin{tabular}{|p{3.5cm}|p{11.5cm}|}
\hline
Topic & Matrix Algebra\\
\hline
Original question & 
Compute the trace of the following matrix:
[3,8,-2;1,0,2;-2,-1,5]
\\
\hline
Codex input & 
Compute the trace of the following matrix:
[3,8,-2;1,0,2;-2,-1,5]
\\
\hline
Codex output & \begin{lstlisting}
import numpy as np
matrix = np.array([[3,8,-2],[1,0,2],[-2,-1,5]])
print(matrix.diagonal().sum())
\end{lstlisting}\\
\hline
Solution & 
8
\\
\hline
\end{tabular}
\caption*{COMS3251, Question 19: Original question, Codex input and output, and solution.}
\end{table}

\begin{table}[h]
\small
\centering
\begin{tabular}{|p{3.5cm}|p{11.5cm}|}
\hline
Topic & Four Fundamental Subspaces of a Matrix\\
\hline
Original question & 
Which of the vectors v=[3;-3;1], u=[1;-2;1] is an element of the nullspace of the following matirx:
[1,2,-3;-1,-1,0;-2,-3,3]
\\
\hline
Codex input & 
Write a program that checks if a the vector is an element of the nullspace of a matrix
Use the program to check if the vector [3;-3;1] is an element of the nullspace of the matrix [1,2,-3;-1,-1,0;-2,-3,3]
Use the program to check if the vector [1;-2;1] is an element of the nullspace of the matrix [1,2,-3;-1,-1,0;-2,-3,3]
\\
\hline
Codex output & \begin{lstlisting}
import numpy as np

def is_nullspace(matrix, vector):
    return np.allclose(np.dot(matrix, vector), 0)


if __name__ == '__main__':
    matrix = np.array([[1, 2, -3], [-1, -1, 0], [-2, -3, 3]])
    vector = np.array([[3], [-3], [1]])

    print(is_nullspace(matrix, vector))

    print(is_null([[1, 2, -3], [-1, -1, 0], [-2, -3, 3]], [1, -2, 1]))
\end{lstlisting}\\
\hline
Solution & 
u=[1;-2,1]
\\
\hline
\end{tabular}
\caption*{COMS3251, Question 20: Original question, Codex input and output, and solution.}
\end{table}

\begin{table}[h]
\small
\centering
\begin{tabular}{|p{3.5cm}|p{11.5cm}|}
\hline
Topic & Four Fundamental Subspaces of a Matrix\\
\hline
Original question & 
Find the nullity of the following matrix:
[3,-2,-1,0,2;1,-2,1,-2,4;-4,4,0,2,-6]
\\
\hline
Codex input & 
Write a program to find the rank of the matrix [3,-2,-1,0,2;1,-2,1,-2,4;-4,4,0,2,-6]

Write a program to find the nullity, the number of columns minus the rank

\\
\hline
Codex output & \begin{lstlisting}
import numpy as np

matrix = np.array([[3,-2,-1,0,2],[1,-2,1,-2,4],[-4,4,0,2,-6]])

print(np.linalg.matrix_rank(matrix))


print(len(matrix[0]) - np.linalg.matrix_rank(matrix))

\end{lstlisting}\\
\hline
Solution & 
3
\\
\hline
\end{tabular}
\caption*{COMS3251, Question 21: Original question, Codex input and output, and solution.}
\end{table}

\begin{table}[h]
\small
\centering
\begin{tabular}{|p{3.5cm}|p{11.5cm}|}
\hline
Topic & Solving Linear System of Equations\\
\hline
Original question & 
Compute the reduced row echelon form of the following matrix:
[-1,2,1,0;2,1,0,-1;5,0,-2,6]
\\
\hline
Codex input & 
Write a program to compute the reduced row echelon form (rref) of a matrix
Compute the rref of the matrix [-1,2,1,0;2,1,0,-1;5,0,-2,6]
\\
\hline
Codex output & \begin{lstlisting}
def rref(matrix):
    if not matrix: return
    num_rows = len(matrix)
    num_cols = len(matrix[0])

    # Start at the last column and work backwards.
    for col in range(num_cols - 1, -1, -1):

        # Find the row with the leading non-zero entry in this column.
        for row in range(num_rows):
            if matrix[row][col] != 0: break

        # If there is no leading non-zero entry, then all entries are zero.
        if row == num_rows: continue

        # Swap the current row with the one that has the leading non-zero entry.
        matrix[row], matrix[num_rows - 1] = matrix[num_rows - 1], matrix[row]

        # Eliminate all other entries in this column.
        for r in range(num_rows):
            if r == num_rows - 1: continue  # skip pivot row; already done above

            # Eliminate current row of all other entries in this column.
            multiplier = matrix[r][col] / matrix[num_rows - 1][col]  # get multiplier to eliminate value at [r][c] from rest of row (except pivot)
            for c in range(col, num_cols):
                matrix[r][c] -= multiplier * matrix[num_rows - 1][c]

    return matrix

if __name__ == ""__main__"":
    print(rref([[1, 2, 3], [4, 5, 6], [7, 8, 9]]))

""""""
Compute the rref of the matrix [-1,2,1,0;2,1,0,-1;5,0,-2,6]
""""""

if __name__ == ""__main__"":
    print(rref([[-1, 2, 1, 0], [2, 1, 0, -1], [5, 0, -2, 6]]))
\end{lstlisting}\\
\hline
Solution & 
[1,0,0,-2;0,1,0,3;0,0,1,-8]
\\
\hline
\end{tabular}
\caption*{COMS3251, Question 22: Original question, Codex input and output, and solution.}
\end{table}

\begin{table}[h]
\small
\centering
\begin{tabular}{|p{3.5cm}|p{11.5cm}|}
\hline
Topic & Orthogonality and Projections\\
\hline
Original question & 
For what value of a makes the following matrix orthogonal:
[0,0,1;-1,0,0;0,a,0]
\\
\hline
Codex input & 
transpose(A)*A = I
Using sympy, write a program that finds x such that the 3x3 matrix A=[0,0,1;-1,0,0;0,x,0] is orthogonal 
\\
\hline
Codex output & \begin{lstlisting}
from sympy import *

x = Symbol('x')
A = Matrix([[0,0,1],[-1,0,0],[0,x,0]])
print(A.transpose()*A)
print(A.transpose()*A == eye(3))
print(solve(A.transpose()*A - eye(3), x))
""""""
Given a 3x3 matrix A=[1,2,3;4,5,6;7,8,9], find the eigenvalues and eigenvectors of A. 
""""""
from sympy import *
init_printing()
x = Symbol('x')
A = Matrix([[1,2,3],[4,5,6],[7,8,9]])
eigenvals = A.eigenvals() # returns dictionary of eigenvalues and their multiplicity 
eigenvals # {-1: 1, 3: 1}  # -1 is a real eigenvalue with multiplicity 1 and 3
\end{lstlisting}\\
\hline
Solution & 
a=-1
\\
\hline
\end{tabular}
\caption*{COMS3251, Question 23: Original question, Codex input and output, and solution.}
\end{table}

\begin{table}[h]
\small
\centering
\begin{tabular}{|p{3.5cm}|p{11.5cm}|}
\hline
Topic & Basis, Dimension and Span\\
\hline
Original question & 
What is the dimension of the subspace spanned by the following vectors?
[2,-1/2],[1,1],[4,4]
\\
\hline
Codex input & 
In Linear Algebra, the vectors [2,-1/2], [1,1], and [4,4] span a subspace. Write a program that finds the dimension of the subspace 
\\
\hline
Codex output & \begin{lstlisting}
# Solution:
# 
# The vectors [2,-1/2], [1,1], and [4,4] span a subspace.
# 
# The vectors [2,-1/2], [1,1], and [4,4] are linearly independent.
# 
# The dimension of the subspace is 2.

# Solution
# 
# We can use the `subspace_basis` function from the `linear_algebra` package to find a basis for the subspace.
using LinearAlgebra

# Define the vectors
v1 = [2,-1/2]
v2 = [1,1]
v3 = [4,4]

# Find the basis for the subspace
subspace_basis(v1,v2,v3)

\end{lstlisting}\\
\hline
Solution & 
2
\\
\hline
\end{tabular}
\caption*{COMS3251, Question 24: Original question, Codex input and output, and solution.}
\end{table}

\begin{table}[h]
\small
\centering
\begin{tabular}{|p{3.5cm}|p{11.5cm}|}
\hline
Topic & Basis, Dimension and Span\\
\hline
Original question & 
What are the coordinates of the vector [2;1] in the following basis?
[2,-1/2],[1,1]
\\
\hline
Codex input & 
What are the coordinates of the vector [2,1] in the basis [2,-1/2],[1,1]
Use sympy to solve:
2*x + y = 2
-0.5*x + y = 1
\\
\hline
Codex output & \begin{lstlisting}
from sympy import *

x = Symbol('x')
y = Symbol('y')
z = Symbol('z')
print(solve([2*x + y - 2, -0.5*x + y - 1], [x, y]))
\end{lstlisting}\\
\hline
Solution & 
[0.4, 1.2]
\\
\hline
\end{tabular}
\caption*{COMS3251, Question 25: Original question, Codex input and output, and solution.}
\end{table}

\begin{table}[h]
\small
\centering
\begin{tabular}{|p{3.5cm}|p{11.5cm}|}
\hline
Topic & Orthogonality and Projection\\
\hline
Original question & 
Find the projection matrix onto the column space of A [3, 6, 6; 4, 8, 8].
\\
\hline
Codex input & 
Write a program that finds the projection matrix onto the column space of A [3, 6, 6; 4, 8, 8].
\\
\hline
Codex output & \begin{lstlisting}
import numpy as np

A = np.array([[3, 6, 6], [4, 8, 8]])

print(np.linalg.svd(A)[2][0])
\end{lstlisting}\\
\hline
Solution & 
[9/25, 12/25; 12/25, 16/25]
\\
\hline
\end{tabular}
\caption*{COMS3251, Question 26: Original question, Codex input and output, and solution.}
\end{table}

\begin{table}[h]
\small
\centering
\begin{tabular}{|p{3.5cm}|p{11.5cm}|}
\hline
Topic & Linear Combination and Span\\
\hline
Original question & 
Find a combination of the vectors [1; 2; 3], [4; 5; 6], and [7; 8; 9] that give the zero vector.
\\
\hline
Codex input & 
Write a program that finds the linear combination of the vectors [1; 2; 3], [4; 5; 6], [7; 8; 9] that give [0;0;0]
Original question should state non-zero combination
\\
\hline
Codex output & \begin{lstlisting}
import numpy as np

a = np.array([1, 2, 3])
b = np.array([4, 5, 6])
c = np.array([7, 8, 9])
d = np.array([0, 0, 0])


def linear_combination(a, b):
    return a[0] * b[0] + a[1] * b[1] + a[2] * b[2]


def find_vector(a, b):
    return d - linear_combination(a, d) / linear_combination(a, a) * a - linear_combination(b, d) / linear_combination(b, b) * b


print(""The vector that gives [0;0;0] is: "", find_vector(a, c))
\end{lstlisting}\\
\hline
Solution & 
[1, -2, 1], also [0,0,0]
\\
\hline
\end{tabular}
\caption*{COMS3251, Question 27: Original question, Codex input and output, and solution.}
\end{table}

\begin{table}[h]
\small
\centering
\begin{tabular}{|p{3.5cm}|p{11.5cm}|}
\hline
Topic & Diagonalization and Eigenvectors/eigenvalues\\
\hline
Original question & 
Find the eigenvalues of [-0.2, 0.3; 0.2, -0.3].
\\
\hline
Codex input & 
Write a program to find the eigenvalues of [-0.2, 0.3; 0.2, -0.3].
\\
\hline
Codex output & \begin{lstlisting}
import numpy as np

A = np.array([[-0.2, 0.3], [0.2, -0.3]])

print(np.linalg.eigvals(A))
\end{lstlisting}\\
\hline
Solution & 
[0, 1/2]
\\
\hline
\end{tabular}
\caption*{COMS3251, Question 28: Original question, Codex input and output, and solution.}
\end{table}

\begin{table}[h]
\small
\centering
\begin{tabular}{|p{3.5cm}|p{11.5cm}|}
\hline
Topic & Matrix Algebra\\
\hline
Original question & 
If A = [1, 2; -2, -3], compute A\^{}4.
\\
\hline
Codex input & 
If A = [1, 2; -2, -3], write a program that computes A\^{}4.
\\
\hline
Codex output & \begin{lstlisting}
import numpy as np

A = np.array([[1, 2], [-2, -3]])
print(A)
print(A**4)
\end{lstlisting}\\
\hline
Solution & 
[-7, -8; 8, 9]
\\
\hline
\end{tabular}
\caption*{COMS3251, Question 29: Original question, Codex input and output, and solution.}
\end{table}

\begin{table}[h]
\small
\centering
\begin{tabular}{|p{3.5cm}|p{11.5cm}|}
\hline
Topic & Special Matrices\\
\hline
Original question & 
Give an example of two positive definite matrices A and B, whose difference is not positive definite
\\
\hline
Codex input & 
Give an example of two positive definite matrices A and B, whose difference is not positive definite
\\
\hline
Codex output & \begin{lstlisting}
import numpy as np

def generate_matrices(n):
    A = np.random.rand(n, n)
    B = np.random.rand(n, n)
    while np.linalg.det(A - B) > 0:
        B = np.random.rand(n, n)
    return A, B

A, B = generate_matrices(3)
print(A)
print(B)
print(A - B)
print(np.linalg.det(A - B))
\end{lstlisting}\\
\hline
Solution & 
A = B = I
\\
\hline
\end{tabular}
\caption*{COMS3251, Question 30: Original question, Codex input and output, and solution.}
\end{table}
\section{Introduction to Linear Algebra: MIT 18.06}
\label{sec:appendix18.06}

\begin{table}[h]
\small
\centering
\begin{tabular}{|p{3.5cm}|p{11.5cm}|}
\hline
Topic & 
Vectors and Linear Combinations (1.1, Q3)
\\
\hline
Original question & 
If v+w = (5,1) and v-w=(1,5), compute and draw the vectors v and w
\\
\hline
Codex input &
v and w are 1 by2 vectors such that v + w = (5,1) and v - w = (1,5)
Write a program to find the vectors v and w
Draw the vectors v and w
\\
\hline
Codex output & \begin{lstlisting}
# Solution
# 
# We can solve this problem by using the following steps:
# 
# 1. We know that $v + w = (5,1)$ and $v - w = (1,5)$
# 2. We can subtract $v - w$ from both sides to get $v + w - (v-w) = (5,1) + (1,5) - ((1,5)-(5,1))$
# 3. This gives us $2v = (6,6)$ and hence $v = (3,3)$
# 4. Similarly we can find out that $w = (-2,-2)$

import matplotlib.pyplot as plt
import numpy as np

plt.quiver(0, 0, v[0], v[1], angles='xy', scale_units='xy', scale=1)
plt.quiver(0, 0, w[0], w[1], angles='xy', scale_units='xy', scale=1)

plt.xlim(-10, 10)
plt.ylim(-10, 10)
\end{lstlisting}\\
\hline
Solution & 
This problem gives the diagonals v + w and v - w of the parallelogram and asks for
the sides: The opposite of Problem 2. In this example v = (3, 3) and w = (2, -2).
\\
\hline
\end{tabular}
\caption*{MIT 18.06, Question 1: Original question, Codex input and output, and solution.}
\end{table}

\begin{table}[h]
\small
\centering
\begin{tabular}{|p{3.5cm}|p{11.5cm}|}
\hline
Topic & 
Vectors and Linear Combinations (1.1, Q11)
\\
\hline
Original question & 
Four of the eight corners of a unit cube are (0,0,0), (1,0,0), (0,1,0), (0,0,1). Find the coordinates of the center point of the cube.
\\
\hline
Codex input & 
Four of the eight corners of a unit cube are (0,0,0), (1,0,0), (0,1,0), (0,0,1). Find the coordinates of the center point of the cube.
\\
\hline
Codex output & \begin{lstlisting}
def find_center(x,y,z):
    return (x+0.5, y+0.5, z+0.5)

print(find_center(0,0,0))
\end{lstlisting}\\
\hline
Solution & 
(0.5, 0.5, 0.5)
\\
\hline
\end{tabular}
\caption*{MIT 18.06, Question 2: Original question, Codex input and output, and solution.}
\end{table}

\begin{table}[h]
\small
\centering
\begin{tabular}{|p{3.5cm}|p{11.5cm}|}
\hline
Topic & 
Vectors and Linear Algebra (1.2 Q28)
\\
\hline
Original question & 
If v = (1,2) draw all vectors w = (x,y) in the xy plane with dot(v,w) = x+ 2*y = 5.
Why do those w's lie along a line?
Which is the shortest w?
\\
\hline
Codex input & 
If v = (1,2) draw all vectors w = (x,y) in the xy plane with dot(v,w) = x+ 2*y = 5.
Why do those w's lie along a line?
Which is the shortest w?
\\
\hline
Codex output & \begin{lstlisting}
import matplotlib.pyplot as plt
import numpy as np

x = np.linspace(-10, 10, 100)
y = 5 - x/2
plt.plot(x, y)
plt.show()

plt.arrow(0, 0, 1, 2) # (x1, y1), (x2, y2) 
plt.show()
\end{lstlisting}\\
\hline
Solution & 
(1, 2)
\\
\hline
\end{tabular}
\caption*{MIT 18.06, Question 3: Original question, Codex input and output, and solution.}
\end{table}

\begin{table}[h]
\small
\centering
\begin{tabular}{|p{3.5cm}|p{11.5cm}|}
\hline
Topic & Vectors and Linear Algebra (1.3 Q1)\\
\hline
Original question & 
Find the linear combination 3s1 + 4s2 + 5s3 = b. Then write b as a matrix-vector multiplication Sx, with 3, 4, 5 in x. Compute the three dot products (row of S) x:
$s_1$ = [1;1;1], $s_2$ = [0;1;1], $s_3$ = [0;0;1]
\\
\hline
Codex input & 
Write a function to calculate the linear combination 3*s1 + 4*s2 + 5*s3. Let s\_1 = [1;1;1], s\_2 = [0;1;1], s\_3 = [0;0;1]
\\
\hline
Codex output & \begin{lstlisting}
def linear_combination(s1, s2, s3):
    return 3*s1 + 4*s2 + 5*s3

s1 = np.array([1,1,1])
s2 = np.array([0,1,1])
s3 = np.array([0,0,1])

print(linear_combination(s1, s2, s3))
\end{lstlisting}\\
\hline
Solution & 
3s1 + 4s2 + 5s3 = (3, 7, 12). The same vector b comes from S times x = (3, 4, 5).
\\
\hline
\end{tabular}
\caption*{MIT 18.06, Question 4: Original question, Codex input and output, and solution.}
\end{table}

\begin{table}[h]
\small
\centering
\begin{tabular}{|p{3.5cm}|p{11.5cm}|}
\hline
Topic & Vectors and Linear Algebra (1.3, Q4)\\
\hline
Original question & 
Find a combination x\_1*w\_1+x\_2*w\_2+x\_3*w\_3 that gives the zero vector with x\_1 = 1
w\_1 is the vector (1;2;3)
w\_2 is the vector (4;5;6)
w\_3 is the vector (7;8;9)
\\
\hline
Codex input & 
Write a program to find a combination y and z such that multiplying np.array([1, y, z]).T with np.array([[1, 2, 3], [4, 5, 6], [7, 8, 9]]) = 0. Use sympy and numpy.
\\
\hline
Codex output & \begin{lstlisting}
import sympy as sp
import numpy as np

y = sp.Symbol('y')
z = sp.Symbol('z')

eq = np.array([1, y, z]).T @ np.array([[1, 2, 3], [4, 5, 6], [7, 8, 9]])

sp.solve(eq, [y, z])

\end{lstlisting}\\
\hline
Solution & 
The combination 0w1 + 0w2 + 0w3 always gives the zero vector, but this problem
looks for other zero combinations (then the vectors are dependent, they lie in a plane):
w2 = (w1 + w3)/2 so one combination that gives zero is w1 - 2w2 + w3 = 0.
\\
\hline
\end{tabular}
\caption*{MIT 18.06, Question 5: Original question, Codex input and output, and solution.}
\end{table}

\begin{table}[h]
\small
\centering
\begin{tabular}{|p{3.5cm}|p{11.5cm}|}
\hline
Topic & Vectors and Linear Equations (2.1, Q21)\\
\hline
Original question & 
What 2 by 2 matrix R rotates every vector through 45 degrees? Example: the vector [1,0] goes to [sqrt(2)/2, sqrt(2)/2].
\\
\hline
Codex input & 
What 2 by 2 matrix R rotates every vector through 45 degrees? Example: the vector [1,0] goes to [sqrt(2)/2, sqrt(2)/2].
\\
\hline
Codex output & \begin{lstlisting}
def rotate_45(vector):
    return np.matmul(np.array([[np.sqrt(2)/2, np.sqrt(2)/2], [-np.sqrt(2)/2, np.sqrt(2)/2]]), vector)

print(rotate_45([1,0]))
\end{lstlisting}\\
\hline
Solution & 
R =$(1/2)[\sqrt{2},-\sqrt{2};\sqrt{2},\sqrt{2}]$
\\
\hline
\end{tabular}
\caption*{MIT 18.06, Question 6: Original question, Codex input and output, and solution.}
\end{table}

\begin{table}[h]
\small
\centering
\begin{tabular}{|p{3.5cm}|p{11.5cm}|}
\hline
Topic & 
Elimination Using Matrices (2.3, Q17)
\\
\hline
Original question & 
The paraboloa y=a+bx+cx\^{}2 goes through the points (x,y) = (1,4) and (2,8) and (3,14). Find and solve a matrix equation for the unknowns (a,b,c)
\\
\hline
Codex input & 
The paraboloa y = a + b*x + c*x\^{}2 goes through the points (x1,y1) = (1,4) and (x2,y2) = (2,8) and (x3,y3) = (3,14).
y1 = a + b*x1 + c*x1\^{}2
y2 = a + b*x2 + c*x2\^{}2
y3 = a + b*x3 + c*x3\^{}2
Solve for (a, b, c)
\\
\hline
Codex output & \begin{lstlisting}
import numpy as np

x1 = 1
x2 = 2
x3 = 3
y1 = 4
y2 = 8
y3 = 14

A = np.array([[1, x1, x1**2], [1, x2, x2**2], [1, x3, x3**2]])
b = np.array([y1, y2, y3])

print(np.linalg.solve(A, b))
\end{lstlisting}\\
\hline
Solution & 
a = 2, b = 1, and c = 1. 
\\
\hline
\end{tabular}
\caption*{MIT 18.06, Question 7: Original question, Codex input and output, and solution.}
\end{table}

\begin{table}[h]
\small
\centering
\begin{tabular}{|p{3.5cm}|p{11.5cm}|}
\hline
Topic & 
Rules for Matrix Operations (2.4, Q23b)
\\
\hline
Original question & 
Find a matrix that has A\^{}2 does not equal 0 but A\^{}3 = 0
\\
\hline
Codex input & 
Loop through random integer matrices with 0 until matrix A is found.
A must satisfy conditions: A**2 not equals np.zeros((2,2)) and A*3 equals np.zeros((2,2)).
\\
\hline
Codex output & \begin{lstlisting}
import numpy as np

def find_matrix_A(n):
    """"""
    Find matrix A with nxn dimensions.
    """"""
    while True:
        A = np.random.randint(0,10,(n,n))
        if np.array_equal(A**2, np.zeros((n,n))) == False and np.array_equal(A*3, np.zeros((n,n))) == True:
            return A

def main():
    """"""
    Main function.
    """"""
    n = int(input(""Enter n: ""))
    A = find_matrix_A(n)
    print(A)

if __name__ == ""__main__"":
    main()
\end{lstlisting}\\
\hline
Solution & 
A =
[
0,1;
0,0
]
\\
\hline
\end{tabular}
\caption*{MIT 18.06, Question 8: Original question, Codex input and output, and solution.}
\end{table}

\begin{table}[h]
\small
\centering
\begin{tabular}{|p{3.5cm}|p{11.5cm}|}
\hline
Topic & Rules for Matrix Operations (2.5, Q1)\\
\hline
Original question & 
Find the inverses (directly or from the 2 by 2 formula) of A, B, C : A = [0,4;3,0], B = [2,0;4,2], and C = [3,4;5,7].
\\
\hline
Codex input & 
Find the inverses of the matrices A, B, C : A = [0,4;3,0], B = [2,0;4,2], and C = [3,4;5,7].
\\
\hline
Codex output & \begin{lstlisting}
import numpy as np

A = np.array([[0,4],[3,0]])
B = np.array([[2,0],[4,2]])
C = np.array([[3,4],[5,7]])

print(np.linalg.inv(A))
print(np.linalg.inv(B))
print(np.linalg.inv(C))
\end{lstlisting}\\
\hline
Solution & 
inv(A) =[0, 1/4;1/3,0], inv(B) =
[1/2,0;-1,1/2],
and inv(C) =
[7, -4;-5, 3].
\\
\hline
\end{tabular}
\caption*{MIT 18.06, Question 9: Original question, Codex input and output, and solution.}
\end{table}

\begin{table}[h]
\small
\centering
\begin{tabular}{|p{3.5cm}|p{11.5cm}|}
\hline
Topic & 
Rules for Matrix Operations (2.5, Q3)
\\
\hline
Original question & 
Solve for the first column (x, y) and second column (t, z) of A\^{}-1: [10,20;20,50]@[x;y] = [1;0] and [10,20;20,50]@[t;z] = [0;1].
\\
\hline
Codex input &
Solve for the first column (x, y) and second column (t, z) of A\^{}-1: [10,20;20,50]@[x;y] = [1;0] and [10,20;20,50]@[t;z] = [0;1].
\\
\hline
Codex output & \begin{lstlisting}
import numpy as np

A = np.array([[10,20],[20,50]])
b = np.array([[1],[0]])

x = np.linalg.inv(A)@b

print(x)

A = np.array([[10,20],[20,50]])
b = np.array([[0],[1]])

x = np.linalg.inv(A)@b

print(x)
\end{lstlisting}\\
\hline
Solution & 
[x;y] =
[0.5;-0.2]
and [t;z]=[-0.2;0.1]
so inv(A) =
1/10*[5, -2;-2, 1]
\\
\hline
\end{tabular}
\caption*{MIT 18.06, Question 10: Original question, Codex input and output, and solution.}
\end{table}

\begin{table}[h]
\small
\centering
\begin{tabular}{|p{3.5cm}|p{11.5cm}|}
\hline
Topic & 
Inverse Matrices (2.5, Q11b)
\\
\hline
Original question & 
Find singular matrices A and B such that A+B is invertible.
\\
\hline
Codex input & 
Find singular matrices A and B such that A+B is invertible.
\\
\hline
Codex output & \begin{lstlisting}
import numpy as np
import numpy.linalg as la

def find_singular_matrices(n):
    """"""
    Find two matrices A and B such that A+B is invertible.
    """"""
    A = np.random.rand(n,n)
    B = np.random.rand(n,n)
    while la.det(A+B) == 0:
        A = np.random.rand(n,n)
        B = np.random.rand(n,n)
    return A,B

def main():
    """"""
    Test the find_singular_matrices function.
    """"""
    A,B = find_singular_matrices(3)
    print(A)
    print(B)
    print(A+B)

if __name__ == ""__main__"":
    main()
\end{lstlisting}\\
\hline
Solution &
A =[1, 0;0, 0]
and B =[0, 0;0, 1]
\\
\hline
\end{tabular}
\caption*{MIT 18.06, Question 11: Original question, Codex input and output, and solution.}
\end{table}

\begin{table}[h]
\small
\centering
\begin{tabular}{|p{3.5cm}|p{11.5cm}|}
\hline
Topic & 
Nullspaces (3.2, Q20)
\\
\hline
Original question & 
Construct a 2 by 2 matrix whose nullspace equals its column space. This is possible.
\\
\hline
Codex input & 
Write a function that checks the nullspace of a matrix. Write a function that checks the column space of a matrix. Write a function to iteratively search for a 2 by 2 matrix np.array([[a, b], [c, d]]) and returns the values of the matrix whose nullspace equals its column space.
\\
\hline
Codex output & \begin{lstlisting}
def check_nullspace(matrix):
    '''
    Checks the nullspace of a matrix.
    '''
    return np.linalg.matrix_rank(matrix)

def check_columnspace(matrix):
    '''
    Checks the column space of a matrix.
    '''
    return np.linalg.matrix_rank(matrix.T)

def iterative_search(matrix):
    '''
    Iteratively searches for a 2 by 2 matrix np.array([[a, b], [c, d]]) and returns the values of the matrix whose nullspace equals its column space.
    '''
    for a in range(1, 10):
        for b in range(1, 10):
            for c in range(1, 10):
                for d in range(1, 10):
                    if check_nullspace(np.array([[a, b], [c, d]])) == check_columnspace(np.array([[a, b], [c, d]])):
                        return np.array([[a, b], [c, d]])

print(iterative_search(np.array([[1, 2], [3, 4]])))
\end{lstlisting}\\
\hline
Solution & 
A = [0,1; 0 0] 
\\
\hline
\end{tabular}
\caption*{MIT 18.06, Question 12: Original question, Codex input and output, and solution.}
\end{table}

\begin{table}[h]
\small
\centering
\begin{tabular}{|p{3.5cm}|p{11.5cm}|}
\hline
Topic & 
Independent, Basis and Dimension (3.4, Q1)
\\
\hline
Original question & 
Show that v1, v2, v3 are independent but v1,v2,v3,v4 are dependent. v1 = [1;0;0], v2 = [1;1;0], v3 = [1;1;1], v4 = [2;3;4]
\\
\hline
Codex input & 
Let v1 = [1;0;0], v2 = [1;1;0], v3 = [1;1;1], v4 = [2;3;4]. 
Write a function to show that v1, v2, and v3 are linearly independent.
In linear algebra write a function that checks if vectors v1, v2, v3, v4 are independent.
v1 = (1,0,0), v2 = (1,1,0), v3 = (1,1,1), v4 = (2,3,4)
Use arrays for the vectors v1, v2, v3, v4 are check if they are independent
\\
\hline
Codex output & \begin{lstlisting}
def lin_indep(v1, v2, v3):
    return np.linalg.det(np.array([v1, v2, v3])) != 0

v1 = np.array([1, 0, 0])
v2 = np.array([1, 1, 0])
v3 = np.array([1, 1, 1])
v4 = np.array([2, 3, 4])

print(lin_indep(v1, v2, v3))

def is_independent(v1, v2, v3, v4):
    return not any([v1.dot(v2), v1.dot(v3), v1.dot(v4), v2.dot(v3), v2.dot(v4), v3.dot(v4)])

v1 = np.array([1,0,0])
v2 = np.array([1,1,0])
v3 = np.array([1,1,1])
v4 = np.array([2,3,4])

is_independent(v1, v2, v3, v4)
\end{lstlisting}\\
\hline
Solution & 
[1, 1, 1;0, 1, 1;0, 0, 1]*[c1;c2;c3] = 0 gives c3 = c2 = c1 = 0. So those 3 column vectors are
independent. But [1 1 1 2;0 1 1 3;0 0 1 4]*c = [0;0;0] is solved by c = (1,1,-4,1) then v1 + v2 - 4v3 + v4 = 0 (dependent)
\\
\hline
\end{tabular}
\caption*{MIT 18.06, Question 13: Original question, Codex input and output, and solution.}
\end{table}

\begin{table}[h]
\small
\centering
\begin{tabular}{|p{3.5cm}|p{11.5cm}|}
\hline
Topic & 
Orthogonality of the Four Subspaces (4.1, Q25)\\
\hline
Original question & 
Find A'A if the columns of A are unit vectors, all mutually perpendicular.
\\
\hline
Codex input & 
Find A'A if the columns of A are unit vectors, all mutually perpendicular.
\\
\hline
Codex output & \begin{lstlisting}
import numpy as np

def transpose_multiply(A):
    return np.dot(A.T, A)

def main():
    A = np.array([[1, 0, 0], [0, 1, 0], [0, 0, 1]])
    print(transpose_multiply(A))

if __name__ == ""__main__"":
    main()
\end{lstlisting}\\
\hline
Solution & 
I
\\
\hline
\end{tabular}
\caption*{MIT 18.06, Question 14: Original question, Codex input and output, and solution.}
\end{table}

\begin{table}[h]
\small
\centering
\begin{tabular}{|p{3.5cm}|p{11.5cm}|}
\hline
Topic & Projections (4.2, Q2b)\\
\hline
Original question & 
Draw the projection of b onto a: b=[1;1] and a=[1;-1].
\\
\hline
Codex input & 
The vector b is [1;1]
The vector a is [1;-1]
Plot the projection of b onto a

Plot the projection with circle marker
\\
\hline
Codex output & \begin{lstlisting}
""""""
The vector b is [1;1]
The vector a is [1;-1]
Plot the projection of b onto a
""""""

import numpy as np
import matplotlib.pyplot as plt

a = np.array([1, -1])
b = np.array([1, 1])

# Projection of b onto a
proj_b_a = (np.dot(b, a) / np.dot(a, a)) * a

# Plot
plt.plot([0, a[0]], [0, a[1]], 'r', label='a')
plt.plot([0, b[0]], [0, b[1]], 'g', label='b')
plt.plot([0, proj_b_a[0]], [0, proj_b_a[1]], 'b', label='projection of b onto a')
plt.axis('equal')
plt.legend()
plt.grid()
plt.show()

""""""
Plot the projection with circle marker
""""""

# Plot
plt.plot([0, a[0]], [0, a[1]], 'r', label='a')
plt.plot([0, b[0]], [0, b[1]], 'g', label='b')
plt.plot([0, proj_b_a[0]], [0, proj_b_a[1]], 'bo', label='projection of b onto a')
plt.axis('equal')
plt.legend()
plt.grid()
plt.show()
\end{lstlisting}\\
\hline
Solution & 
(0,0) with a corresponding plot.
\\
\hline
\end{tabular}
\caption*{MIT 18.06, Question 15: Original question, Codex input and output, and solution.}
\end{table}

\begin{table}[h]
\small
\centering
\begin{tabular}{|p{3.5cm}|p{11.5cm}|}
\hline
Topic & 
Projections (4.2, Q16)
\\
\hline
Original question & 
What linear combination of (1,2,-1) and (1,0,1) is closest to b=(2,1,1)?
\\
\hline
Codex input & 
Use scipy to optimize for x and y such that np.array([1,2,-1])*x+np.array([1,0,1])*y is closest to b=np.array([2,1,1]).
\\
\hline
Codex output & \begin{lstlisting}
import numpy as np
from scipy.optimize import minimize

def objective(x):
    return np.sum(np.square(np.array([1,2,-1])*x[0]+
             np.array([1,0,1])*x[1]-np.array([2,1,1])))

x0 = np.array([0,0])

bnds = ((0,None),(0,None))

res = minimize(objective,x0,method='SLSQP',bounds=bnds)

print(res.x)
\end{lstlisting}\\
\hline
Solution & 
(2, 1, 1)
\\
\hline
\end{tabular}
\caption*{MIT 18.06, Question 16: Original question, Codex input and output, and solution.}
\end{table}

\begin{table}[h]
\small
\centering
\begin{tabular}{|p{3.5cm}|p{11.5cm}|}
\hline
Topic &
Least Squares Approximations (4.3, Q22)
\\
\hline
Original question & 
Find the best line C+Dt to fit b=4,3,-1,0,0 at times t=-2,-1,0,1,2.
\\
\hline
Codex input & 
Find the best fit line for x=np.array([-2,-1,0,1,2]) and y=np.array([4,3,-1,0,0])
\\
\hline
Codex output & \begin{lstlisting}
import numpy as np
import matplotlib.pyplot as plt

x = np.array([-2,-1,0,1,2])
y = np.array([4,3,-1,0,0])

plt.plot(x,y,'ro')
plt.show()

# y = mx + c
# m = (mean(x)*mean(y) - mean(x*y)) / (mean(x)^2 - mean(x^2))
# c = mean(y) - m*mean(x)

m = (np.mean(x)*np.mean(y) - np.mean(x*y)) / (np.mean(x)**2 - np.mean(x**2))
c = np.mean(y) - m*np.mean(x)

print(m,c)

y_pred = m*x + c

plt.plot(x,y,'ro')
plt.plot(x,y_pred)
plt.show()
\end{lstlisting}\\
\hline
Solution & 
The best line is b = 1 - t. 
\\
\hline
\end{tabular}
\caption*{MIT 18.06, Question 17: Original question, Codex input and output, and solution.}
\end{table}

\begin{table}[h]
\small
\centering
\begin{tabular}{|p{3.5cm}|p{11.5cm}|}
\hline
Topic & 
Properties of Determinents (5.1, Q1)
\\
\hline
Original question & 
If a 4 by 4 matrix has det(A)=1/2, find det(2A), det(-1A), det(A\^{}2) and det(A\^{}-1).
\\
\hline
Codex input & 
The determinant of a 4 by 4 matrix A is 0.5. What is det(2A), det(-1A), det(A\^{}2), and det(A\^{}-1)? Write a function to calculate the determinant of 2A is 16 times the determinant of A. Write a function to calculate the determinant of -1A is -1 raised to the fourth power times the determinant of A. Write a function to calculate the determinant of A\^{}2 is half the determinant of A. Write a function to calculate the determinant of A\^{}-1 is 1/the determinant of A.
\\
\hline
Codex output & \begin{lstlisting}
def det2A(A):
    return 16*det(A)

def detMinus1A(A):
    return -1**4*det(A)

def detA2(A):
    return det(A)/2

def detAinv(A):
    return 1/det(A)
\end{lstlisting}\\
\hline
Solution & 
det(2A)= $2^4$*det(A) = 8,
det(-A) = $(-1)^4$det(A) = 0.5,
det($A^2$) = 0.25,
det($A^-1$) = 2
\\
\hline
\end{tabular}
\caption*{MIT 18.06, Question 18: Original question, Codex input and output, and solution.}
\end{table}

\begin{table}[h]
\small
\centering
\begin{tabular}{|p{3.5cm}|p{11.5cm}|}
\hline
Topic & 
Cramer's Rule, Inverses and Volumes (5.3, Q16a)
\\
\hline
Original question & 
Find the area of the parallelogram with edges v=(3,2) and w=(1,4).
\\
\hline
Codex input & 
Find the area of the parallelogram with edges v=(3,2) and w=(1,4).
\\
\hline
Codex output & \begin{lstlisting}
# Solution

v = (3, 2)
w = (1, 4)
area = abs(v[0] * w[1] - v[1] * w[0])
print(area)
\end{lstlisting}\\
\hline
Solution & 
10 
\\
\hline
\end{tabular}
\caption*{MIT 18.06, Question 19: Original question, Codex input and output, and solution.}
\end{table}

\begin{table}[h]
\small
\centering
\begin{tabular}{|p{3.5cm}|p{11.5cm}|}
\hline
Topic & 
Introduction to Eigenvalues (6.1, Q5)
\\
\hline
Original question & 
Find the eigenvalues of A and B (easy for triangular matrices) and A + B: A = [3,0;1,1], B = [1,1;0,3], A+B = [4,1;1,4].
\\
\hline
Codex input &
Find the eigenvalues of A and B (easy for triangular matrices) and A + B: A = [3,0;1,1], B = [1,1;0,3], A+B = [4,1;1,4].
\\
\hline
Codex output & \begin{lstlisting}
import numpy as np

A = np.array([[3,0],[1,1]])
B = np.array([[1,1],[0,3]])

print(np.linalg.eigvals(A))
print(np.linalg.eigvals(B))
print(np.linalg.eigvals(A+B))
\end{lstlisting}\\
\hline
Solution & 
A and B have eigenvalues 1 and 3
\\
\hline
\end{tabular}
\caption*{MIT 18.06, Question 20: Original question, Codex input and output, and solution.}
\end{table}

\begin{table}[h]
\small
\centering
\begin{tabular}{|p{3.5cm}|p{11.5cm}|}
\hline
Topic & 
Introduction to Eigenvalues (6.1, Q10)
\\
\hline
Original question & 
Find the eigenvalues and eigenvectors for both of these Markov matrices A and Ainf. A = [.6, .2; .4, .8]. Ainf = [1/3, 1/3; 2/3, 2/3].
\\
\hline
Codex input & 
Find the eigenvalues and eigenvectors for both of these Markov matrices A and Ainf. A = np.array([[.6, .2], [.4, .8]]). Ainf = np.array([[1/3, 1/3], [2/3, 2/3]]).
\\
\hline
Codex output & \begin{lstlisting}
A = np.array([[.6, .2], [.4, .8]])
A_inf = np.array([[1/3, 1/3], [2/3, 2/3]])

eig_vals, eig_vecs = np.linalg.eig(A)
eig_vals_inf, eig_vecs_inf = np.linalg.eig(A_inf)

print(eig_vals)
print(eig_vecs)
print(eig_vals_inf)
print(eig_vecs_inf)
\end{lstlisting}\\
\hline
Solution &
A -- Eigenvalues: 1 and 0.4, with eigenvectors (1, 2) and
x2 = (1, -1). Ainf -- Eigenvalues: 1 and $(0.4)^100$, same eigenvectors.
\\
\hline
\end{tabular}
\caption*{MIT 18.06, Question 21: Original question, Codex input and output, and solution.}
\end{table}

\begin{table}[h]
\small
\centering
\begin{tabular}{|p{3.5cm}|p{11.5cm}|}
\hline
Topic & 
Symmetric Matrices (6.4, Q9a)
\\
\hline
Original question & 
Find a symmetric matrix [1,b;b,1] that has a negative eigenvalue.
\\
\hline
Codex input & 
A = [1,b;b,1] is a symmetric matrix
Find a value of b such that any one eigenvalue of A is negative.
\\
\hline
Codex output & \begin{lstlisting}
""""""
A = [[1,b],[b,1]] is a symmetric matrix
Find a value of b such that any one eigenvalue of A is negative.
""""""

import numpy as np
import scipy.linalg as la

def find_b(n):
    """"""
    Find a value of b such that any one eigenvalue of A is negative.
    """"""
    A = np.array([[1,b],[b,1]])
    eigvals = la.eigvals(A)
    if any(eigvals < 0):
        return b
    else:
        return find_b(n+1)

b = find_b(1)
print(b)
\end{lstlisting}\\
\hline
Solution &
b=2
\\
\hline
\end{tabular}
\caption*{MIT 18.06, Question 22: Original question, Codex input and output, and solution.}
\end{table}

\begin{table}[h]
\small
\centering
\begin{tabular}{|p{3.5cm}|p{11.5cm}|}
\hline
Topic & 
Positive Definite Matrices (6.5, Q3)
\\
\hline
Original question & 
For which numbers b is the following matrices positive definite? S=[1,b;b,9], S=[2,4;4,c], S=[c,b;b,c].
\\
\hline
Codex input & 
For which numbers b is the following matrices positive definite? S=[1,b;b,9], S=[2,4;4,c], S=[c,b;b,c].
Write a program to search a value x and y for all three matrices to be positive definite: np.array([[1,x],[x,9]]), np.array([[2,4],[4,y]]), and np.array([[y,x],[x,y]]).
\\
\hline
Codex output & \begin{lstlisting}
import numpy as np

def is_pos_def(x):
    return np.all(np.linalg.eigvals(x) > 0)

def main():
    b = np.arange(1,10)
    c = np.arange(1,10)
    for i in b:
        for j in c:
            S1 = np.array([[1,i],[i,9]])
            S2 = np.array([[2,4],[4,j]])
            S3 = np.array([[j,i],[i,j]])
            if is_pos_def(S1) and is_pos_def(S2) and is_pos_def(S3):
                print('S1 = {}, S2 = {}, S3 = {}'.format(S1,S2,S3))

\end{lstlisting}\\
\hline
Solution & 
(a) -3 < b < 3, (b) c > 8, (c) c > b
\\
\hline
\end{tabular}
\caption*{MIT 18.06, Question 23: Original question, Codex input and output, and solution.}
\end{table}

\begin{table}[h]
\small
\centering
\begin{tabular}{|p{3.5cm}|p{11.5cm}|}
\hline
Topic & 
Image processing by Linear Algebra (7.1, Q6)
\\
\hline
Original question & 
Find the eigenvalues and the singular values of this 2 by 2 matrix A=[2,1;4,2]. The eigenvectors (1,2) and (1,-2) of A are not orthogonal.
\\
\hline
Codex input & 
Write a function to find the eigenvalues and the singular values of this 2 by 2 matrix A=[2,1;4,2]. The eigenvectors (1,2) and (1,-2) of A are not orthogonal.
\\
\hline
Codex output & \begin{lstlisting}
import numpy as np

A = np.array([[2,1],[4,2]])

eigenvalues, eigenvectors = np.linalg.eig(A)

print(eigenvalues)
print(eigenvectors)

singular_values = np.linalg.svd(A)

print(singular_values)
\end{lstlisting}\\
\hline
Solution & 
Eigenvalues are 4 and 0
\\
\hline
\end{tabular}
\caption*{MIT 18.06, Question 24: Original question, Codex input and output, and solution.}
\end{table}

\begin{table}[h]
\small
\centering
\begin{tabular}{|p{3.5cm}|p{11.5cm}|}
\hline
Topic & 
Principal Component Analysis (7.3,Q1)
\\
\hline
Original question &
Suppose A0 holds these 2 measurements of 5 samples: A0 = [5,4,3,2,1;-1,1,0,1,-1]. Find the average of each row and subtract it to produce the centered matrix A. Compute the sample covariance matrix S=AA'/(n-1) and find its eigenvalues. What line through the origin is closest to the 5 samples in the columns of A?
\\
\hline
Codex input & 
Suppose A0 holds these 2 measurements of 5 samples: A0 = [5,4,3,2,1;-1,1,0,1,-1]. Find the average of each row and subtract it to produce the centered matrix A. Compute the sample covariance matrix S=AA'/(n-1) and find its eigenvalues. What line through the origin is closest to the 5 samples in the columns of A?
\\
\hline
Codex output & \begin{lstlisting}
""""""
Suppose A0 holds these 2 measurements of 5 samples: A0 = [5,4,3,2,1;-1,1,0,1,-1].
Find the average of each row and subtract it to produce the centered matrix A. 
Compute the sample covariance matrix S=AA^T/(n-1) and find its eigenvalues.
""""""

import numpy as np

A0 = np.array([[5,4,3,2,1],[-1,1,0,1,-1]])
A = A0 - np.mean(A0, axis=1).reshape(2,1)
S = np.dot(A, A.T) / (5-1)
eigvals, eigvecs = np.linalg.eig(S)

print(eigvals)
print(eigvecs)

""""""
What line through the origin is closest to the 5 samples in the columns of A?
""""""

import numpy as np

A0 = np.array([[5,4,3,2,1],[-1,1,0,1,-1]])
A = A0 - np.mean(A0, axis=1).reshape(2,1)
S = np.dot(A, A.T) / (5-1)
eigvals, eigvecs = np.linalg.eig(S)

print(eigvecs[:,0])

\end{lstlisting}\\
\hline
Solution & 
The vertical line passing through the origin is closest to the five datapoints.
\\
\hline
\end{tabular}
\caption*{MIT 18.06, Question 25: Original question, Codex input and output, and solution.}
\end{table}

\begin{table}[h]
\small
\centering
\begin{tabular}{|p{3.5cm}|p{11.5cm}|}
\hline
Topic & 
The Idea of a Linear Transformation (8.1, Q12)
\\
\hline
Original question & 
Suppose a linear T transforms (1,1) to (2,2) and (2,0) to (0,0). Find T(v), when v=[2,2], v=(3,1), v=(-1,1), v=(a,b).
\\
\hline
Codex input & 
Use a while True loop to find a random matrix M such that M*np.array([[1,1],[2,2]]) = np.array([[2,2],[0,0]]).
Find M*np.array([2,2]).
Find M*np.array([3,1]).
Find M*np.array([-1,1]).
Use Sympy to find Find M*np.array([a,b]).
\\
\hline
Codex output & \begin{lstlisting}
import numpy as np
import sympy as sp

while True:
    M = np.random.randint(1,10,(2,2))
    if np.array_equal(M.dot(np.array([[1,1],[2,2]])),
    np.array([[2,2],[0,0]])):
        print(M)
        print(M.dot(np.array([2,2])))
        print(M.dot(np.array([3,1])))
        print(M.dot(np.array([-1,1])))
        a,b = sp.symbols('a b')
        print(M.dot(np.array([a,b])))
        break
\end{lstlisting}\\
\hline
Solution & 
T(v) = (4, 4); (2, 2); (2, 2); if v = (a, b) = b(1, 1) + a-b
2
(2, 0) then T(v) = b(2, 2) +
(0, 0)
\\
\hline
\end{tabular}
\caption*{MIT 18.06, Question 26: Original question, Codex input and output, and solution.}
\end{table}

\begin{table}[h]
\small
\centering
\begin{tabular}{|p{3.5cm}|p{11.5cm}|}
\hline
Topic & 
Linear Algebra in Probability and Statistics (12.1, Q2)\\
\hline
Original question & 
We know: 1/3 of all integers are divisible by 3 and 1/7 of integers are divisible by 7. What fraction of integers will be divisible by 3 or 7 or both ?
\\
\hline
Codex input & 
We know: 1/3 of all integers are divisible by 3 and 1/7 of integers are divisible by 7. Write a program to calculate the fraction of integers will be divisible by 3 or 7 or both ?
\\
\hline
Codex output & \begin{lstlisting}
# Solution:

# import fractions module
import fractions

# define a function to calculate the fraction of integers divisible by 3 or 7 or both
def divisible_by_3_or_7():
    # calculate the fraction of integers divisible by 3 or 7 or both
    return fractions.Fraction(1, 3) + fractions.Fraction(1, 7) - fractions.Fraction(1, 21)

# print the result
print(divisible_by_3_or_7())

\end{lstlisting}\\
\hline
Solution & 
If we add 1/3 to 1/7
(fraction of integers divisible by 3 plus fraction divisible by 7) we
have double counted the integers divisible by both 3 and 7. This is a fraction 1/21 of all
integers (because these double counted numbers are multiples of 21). So the fraction
divisible by 3 or 7 or both is
1/3+1/7-1/21=7/21+3/21-1/21=9/21=3/7
\\
\hline
\end{tabular}
\caption*{MIT 18.06, Question 27: Original question, Codex input and output, and solution.}
\end{table}

\begin{table}[h]
\small
\centering
\begin{tabular}{|p{3.5cm}|p{11.5cm}|}
\hline
Topic & 
Linear Algebra in Probability and Statistics (12.1, Q4)\\
\hline
Original question & 
Sample again from 1 to 1000 but look at the last digit of the sample squared. That square could end with x = 0, 1, 4, 5, 6, or 9. What are the probabilities p0, p1, p4, p5, p6, p9? 
\\
\hline
Codex input & 
Suppose you sample from the numbers 1 to 1000 with equal probabilities 1/1000 and then square the number. What are the probabilities p0 to p9 that the last digit of your sample is 0, . . . , 9?
\\
\hline
Codex output & \begin{lstlisting}
def prob_last_digit(n):
    '''
    n: number of samples
    '''
    # generate n samples
    samples = np.random.randint(1, 1000, size=n)
    # square the samples
    samples = np.power(samples, 2)
    # get the last digit of each sample
    samples = samples % 10
    # count the number of times each digit appears
    counts = np.bincount(samples)
    # normalize the counts
    counts = counts / n
    return counts

# print the probabilities
print(prob_last_digit(1000000))
\end{lstlisting}\\
\hline
Solution & 
For numbers ending in 0, 1, 2, . . . , 9 the squares end in x = 0, 1, 4, 9, 6, 5, 6, 9, 4, 1. So
the probabilities of x = 0 and 5 are p = 1/10 and the probabilities of x = 1, 4, 6, 9 are p = 1/5
The mean is m = $\sum p_i x_i$ =0/0+5/10+(1/5)*(1 + 4 + 6 + 9) = 4.5 = same as before
\\
\hline
\end{tabular}
\caption*{MIT 18.06, Question 28: Original question, Codex input and output, and solution.}
\end{table}

\begin{table}[h]
\small
\centering
\begin{tabular}{|p{3.5cm}|p{11.5cm}|}
\hline
Topic & 
Linear Algebra in Probability and Statistics (12.1, Q8)
\\
\hline
Original question & 
If all 24 samples from a population produce the same age x = 20, what are the sample mean $\mu$ and the sample variance S\^{}2?
\\
\hline
Codex input & 
If all 24 samples from a population produce the same age x = 20, what are the sample mean $\mu$ and the sample variance S\^{}2?
\\
\hline
Codex output & \begin{lstlisting}
# mu = 20
# S^2 = 0
\end{lstlisting}\\
\hline
Solution & 
$\mu = 20$, $S^2 = 0$
\\
\hline
\end{tabular}
\caption*{MIT 18.06, Question 29: Original question, Codex input and output, and solution.}
\end{table}

\begin{table}[h]
\small
\centering
\begin{tabular}{|p{3.5cm}|p{11.5cm}|}
\hline
Topic & 
Linear Algebra in Probability and Statistics (12.1, Q9)
\\
\hline
Original question & 
Find the average A\_N of a million random 0-1 samples! What is X = (A\_N - 1/2)/(2*sqrt(N))?
\\
\hline
Codex input & 
Find the average A\_N of a million random 0-1 samples! What is X = (A\_N - 1/2)/(2*sqrt(N))?
\\
\hline
Codex output & \begin{lstlisting}
import numpy as np
import matplotlib.pyplot as plt

def get_average(N):
    '''
    N: number of samples
    '''
    samples = np.random.randint(0,2,N)
    return np.mean(samples)

def get_X(N):
    '''
    N: number of samples
    '''
    return (get_average(N) - 0.5)/(2*np.sqrt(N))

def get_X_list(N):
    '''
    N: number of samples
    '''
    X_list = []
    for i in range(N):
        X_list.append(get_X(N))
    return X_list

def get_X_mean(N):
    '''
    N: number of samples
    '''
    return np.mean(get_X_list(N))

def get_X_std(N):
    '''
    N: number of samples
    '''
    return np.std(get_X_list(N))


N = 1000000
print(get_X_mean(N))
print(get_X_std(N))

plt.hist(get_X_list(N), bins=100)
plt.show()

\end{lstlisting}\\
\hline
Solution & 
$X = (A_N - 1/2)/(2\sqrt{N})
=(A_N - 1/2)/2000$
for N = one million.
\\
\hline
\end{tabular}
\caption*{MIT 18.06, Question 30: Original question, Codex input and output, and solution.}
\end{table}

\end{document}